\documentclass{article}
\usepackage[final,main]{neurips_2026}

\usepackage[utf8]{inputenc}
\usepackage[T1]{fontenc}
\usepackage{hyperref}
\usepackage{url}
\usepackage{amsmath, amssymb, amsthm, amsfonts}
\usepackage{microtype}
\usepackage{bm}
\usepackage{xcolor}
\usepackage{algorithm, algorithmic}
\usepackage{graphicx}
\usepackage{booktabs}
\usepackage{array}
\usepackage{multirow}
\usepackage{caption}
\usepackage{subcaption}
\usepackage{float}
\definecolor{linknavy}{RGB}{0,62,126}
\hypersetup{colorlinks=true, linkcolor=linknavy, citecolor=linknavy, urlcolor=linknavy}

\theoremstyle{plain}
\newtheorem{theorem}{Theorem}
\newtheorem{proposition}{Proposition}
\newtheorem{corollary}{Corollary}
\newtheorem{lemma}{Lemma}

\theoremstyle{definition}
\newtheorem{definition}{Definition}

\theoremstyle{remark}

\title{Coupling-Aware Reinforcement Learning \\for Co-Evolving Graph Games}

\newlength{\authnamew}
\author{%
  \settowidth{\authnamew}{Benoit Montreuil}%
  \makebox[\authnamew][c]{Mina Kim} \quad
  \makebox[\authnamew][c]{Guanghui Lan} \quad
  \makebox[\authnamew][c]{Benoit Montreuil} \\
  H.\ Milton Stewart School of Industrial and Systems Engineering \\
  Georgia Institute of Technology \\
  Atlanta, GA 30332 \\
  \texttt{mkim939@gatech.edu} \quad \texttt{george.lan@isye.gatech.edu} \\
  \texttt{benoit.montreuil@isye.gatech.edu}
}

\begin{document}
\maketitle

\begin{abstract}
Building hydrogen station networks, electric vehicle charging grids, or vaccine distribution systems requires coordinating two separately operated graphs, demand-side facilities and the supply-side networks they depend on, where each operator's payoff depends on the other's choices.
We formalize this as the Co-Evolving Graph Game (CEGG), a setting that sits between facility location, interdependent-network analysis, and graph-based multi-agent reinforcement learning (MARL) but is not jointly addressed by any of them.
We propose Cross-Graph Attention with Policy Mirror Descent (CGA-PMD), pairing cross-graph attention, which lets each agent read only the partner-graph nodes structurally coupled to its own, with a policy mirror descent update for stable training.
Across three reward domains and a range of practical network scales, we find that the structure of the coupling selects which method wins: at small scale most methods are competitive, but as scale grows or coupling becomes irregular, CGA-PMD is the one method that stays reliable: it attains the highest success count on each of five real city networks, while own-only, unmasked-attention, and Proximal Policy Optimization (PPO)-based variants degrade in distinct ways.
Our theoretical analysis addresses the optimizer side: the Kullback-Leibler (KL)-regularized update admits an iteration-uniform cumulative drift bound, whereas the clip update admits no such bound on the iterations where the clip does not bind.
The implementation is available at \url{https://github.com/Mina-Kim-Github/cegg-cga-pmd}.
\end{abstract}

\section{Introduction}

\begin{figure}[t]
\centering
\includegraphics[width=0.88\textwidth]{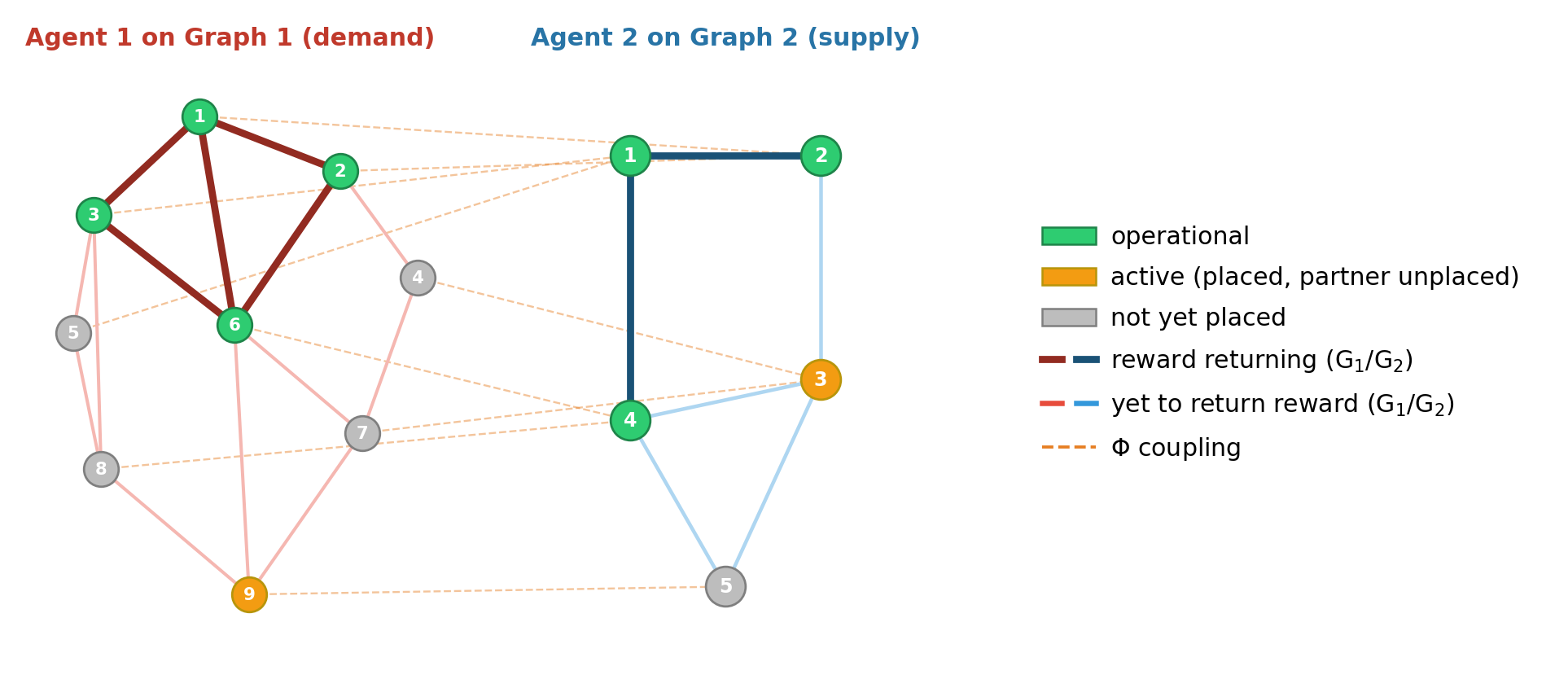}
\caption{CEGG setup: two agents expand separate graphs.}
\label{fig:cegg_concept}
\end{figure}

Consider two agents that must sequentially expand separate graphs (Figure~\ref{fig:cegg_concept}): Agent~1 adds nodes to graph $G_1$ and Agent~2 to graph $G_2$, and a node on $G_1$ becomes operational, starting a diffusion process on its neighboring edges, only when a coupling-partner node on $G_2$ has also been activated.
This structure arises across many infrastructure-expansion problems, hydrogen refueling stations need reachable production plants, electric vehicle (EV) chargers need grid capacity, vaccination sites need cold-chain logistics, 5G towers need fiber backhaul, so each agent's optimal placement depends on the other's choices, yet the two agents act on different graphs.
We use hydrogen as the primary case study but the formulation is domain-agnostic.

Existing methods do not directly address this setting.
Facility location operates on a single network topology, not on two separate coupled graphs expanded in tandem.
Interdependent-network analysis studies how fixed coupled networks fail under perturbations, not how multiple decision-makers should expand them.
Standard multi-agent reinforcement learning (MARL) assumes either a shared observation space or full decentralization, neither of which captures the structured coupling across distinct graphs.
Practitioners therefore resort to ad-hoc heuristics or treat one network as fixed while optimizing the other. 
This paper closes this gap with the following contributions:
\begin{enumerate}
    \item \textbf{Problem formalization.} We define the Co-Evolving Graph Game (CEGG), a class of two-agent sequential network-expansion problems sitting in the gap between facility location, interdependent-network analysis, and graph-MARL.
    \item \textbf{Algorithmic proposal and empirical regime.} We propose CGA-PMD, pairing Cross-Graph Attention with a Policy Mirror Descent update, and benchmark it against five primary baselines covering own-only, coupling-masked, and unmasked-attention architectures crossed with Proximal Policy Optimization (PPO) and Policy Mirror Descent (PMD) optimizers, across three reward calibrations and a range of practical network scales.
    CEGG instances partition into difficulty zones in which different inductive biases win, and CGA-PMD is the method that stays reliable at large scale and under irregular coupling.
    \item \textbf{Theoretical analysis.} We prove an iteration-uniform cumulative drift bound for the idealized KL-regularized update under noisy cross-graph advantages, and the complementary negative statement for the clip update on the iterations where the clip does not bind.
    A neural-architecture extension ties the bound to inter-graph coupling cardinality, and a bias--variance decomposition of the masked context, measured on trained models, says when the $\Phi$ mask helps.
\end{enumerate}

\section{Related Work}
\label{sec:related}

The CEGG sits in a gap among coupled-network analysis, sequential network design, and graph-MARL (Table~\ref{tab:related_compact}): no prior setting captures two separately evolving networks, each expanded by its own agent, where each agent's reward depends on the other network's state.
This prevents direct application of interdependent-network analysis~\citep{artime2024robustness, buldyrev2010catastrophic} (no agency), \citet{mei2026multi}'s heterogeneous-attention MARL on fixed coupled networks, whose attention we adapt with a coupling-aware mask as CGA and use unmasked as the MAPPO-Het baseline, facility-location and reinforcement learning for combinatorial optimization on a single graph~\citep{khalil2017learning, darvariu2024graph, golovin2011adaptive, guo2025unified, laporte2019location}, including RL siting of refueling infrastructure~\citep{kim2026rlhrs}, or graph-MARL with shared observation~\citep{dewitt2020independent, jiang2020graph, lowe2017multi, rashid2020monotonic, sintes2025cognac, yu2022surprising}.
Cross-graph attention with a coupling-aware mask is our architectural response.

\begin{table}[h]
\centering
\caption{Position of the CEGG among adjacent literatures.}
\label{tab:related_compact}
\small
\begin{tabular}{@{}>{\raggedright\arraybackslash}p{0.230\textwidth}>{\raggedright\arraybackslash}p{0.370\textwidth}>{\raggedright\arraybackslash}p{0.150\textwidth}>{\raggedright\arraybackslash}p{0.159\textwidth}@{}}
\toprule
\textbf{Literature class} & \textbf{Representative work} & \textbf{Graph} & \textbf{Agent} \\
\midrule
Graph-MARL on shared graph & 
COGNAC~\citep{sintes2025cognac},
DGN~\citep{jiang2020graph},
HAPPO~\citep{kuba2022happo},
IPPO~\citep{dewitt2020independent},
MADDPG~\citep{lowe2017multi},
MAPPO~\citep{yu2022surprising},
QMIX~\citep{rashid2020monotonic} & 1, fixed & Multi-agent 
\\[3pt]
\midrule
Facility location / RL for graph combinatorial optimization & PPO-Swap~\citep{guo2025unified, laporte2019location},
S2V-DQN~\citep{khalil2017learning, darvariu2024graph, golovin2011adaptive} & 1, expanding & Single agent \\[3pt]
\midrule
Interdependent-network analysis & \citep{artime2024robustness, buldyrev2010catastrophic} & 2, fixed & No agency \\[3pt]
\midrule
Coupled-graph MARL & MAPPO-Het~\citep{mei2026multi} & 2, fixed & Multi-agent \\[3pt]
\midrule
\textbf{CEGG (this work)} & \textbf{CGA-PMD} & \textbf{2, expanding} & \textbf{Multi-agent} \\
\bottomrule
\end{tabular}
\end{table}

CGA-PMD anchors on policy mirror descent (PMD)~\citep{lan2023policy} and its extensions~\citep{li2025policy, protopapas2024policy}, and implements the update in the dual-averaging form of the policy dual averaging (PDA) family~\citep{gao2026actor, ju2024policy}; Mirror Descent Policy Optimization (MDPO)~\citep{tomar2020mirror} is the deep-RL line it draws on.
The policy-gradient convergence literature~\citep{agarwal2021theory, cen2022fast, li2025policy, mei2020global} provides the broader context for our drift analysis in Section~\ref{sec:theory}.

\section{Framework and Algorithm}
\label{sec:formulation}

\subsection{Co-Evolving Graph Games}

\begin{definition}[Co-Evolving Graph Game]
\label{def:cegg}
A Co-Evolving Graph Game (CEGG) is a tuple $\mathcal{G} = (G_1, G_2, \mathcal{S}, \mathcal{A}_1, \mathcal{A}_2, P, R_1, R_2, \Phi, T, \gamma)$:
\begin{itemize}
    \item $G_i = (V_i, E_i)$: Agent $i$'s network, $|V_i| = N_i$, for $i \in \{1,2\}$.
    \item $\mathcal{S} = \{0,1\}^{N_1} \times \{0,1\}^{N_2} \times \mathbb{R}_{\geq 0}^{|E_1|}$: joint state comprising the installation masks $(\mathbf{m}^1_t, \mathbf{m}^2_t)$, with $m_{t,v}^i = 1$ indicating that Agent $i$ has placed node $v$ by time $t$, and a per-edge diffusion state $\mathbf{f}_t$ on $G_1$, where $f_t^{(v,w)} \geq 0$ is the accumulated adoption level on edge $(v,w) \in E_1$.
    \item $\mathcal{A}_i(s_t) = \{v \in V_i : m_{t,v}^i = 0\} \cup \{\emptyset\}$: at each step, Agent $i$ places a single not-yet-installed node or waits (the $\emptyset$ action), updating $\mathbf{m}^i$. We write $a_t^i \in \mathcal{A}_i(s_t)$ for Agent $i$'s action at step $t$; the installation cost is charged only when $a_t^i \neq \emptyset$, i.e.\ when the agent places rather than waits.
    \item $\Phi(v) \subseteq V_2$ for each $v \in V_1$: inter-network coupling assigning each demand node its set of coupling-partner supply nodes. A node $v \in V_1$ becomes operational iff $m^1_{t,v} = 1$ and $\exists u \in \Phi(v)$ with $m^2_{t,u} = 1$. Our experiments use a distance-based form $\Phi(v) = \{u \in V_2 : d_{G_1}(v, u) \leq d_{\max}\}$. We write $\bar{|\Phi|} := \frac{1}{N_1}\sum_{v \in V_1} |\Phi(v)|$ for the mean coupling cardinality (the average number of supply partners per demand node).
    \item $P$: Diffusion dynamics on $G_1$. For each edge $(v,w) \in E_1$ with both endpoints operational:
    \[
    f_{t+1}^{(v,w)} = f_t^{(v,w)} + \!\left(p + q \frac{f_t^{(v,w)}}{C_t^{(v,w)}}\right)\!\left(C_t^{(v,w)} - f_t^{(v,w)}\right)
    \]
    where $p, q$ are innovation and imitation rates and $C_t$ is edge capacity (Bass model~\citep{bass1969new}). This endpoint-activation form follows the freight-energy adoption model of~\citet{kim2026structure}, extended here so that operability depends on a second, separately operated graph.
    \item $R_i$: Agent $i$'s reward (instance-specific; reward instantiation below).
    \item $T \in \mathbb{Z}_{>0}$: episode horizon. $\gamma \in [0, 1)$: discount factor.
\end{itemize}
\end{definition}

The graphs are deliberately asymmetric: adoption, and therefore revenue, occurs on the demand side, so the state carries a diffusion process on $G_1$ only, while installation cost falls on both, and $G_2$ gates which demand nodes are operational. Co-evolving accordingly means co-expansion, both graphs grown by their own operator over the horizon, the axis on which Table~\ref{tab:related_compact} separates the CEGG.

The framework is agnostic to the specific forms of $\Phi$, $P$, and $R$: any coupling relation, state-transition kernel, and instance-specific reward functions are admissible.

We use the word ``game'' in its descriptive sense: two agents with their own rewards, acting on different graphs, coupled through $\Phi$.
What we analyze is the learning process, a training-stability bound and a comparison of learned-policy quality and training reliability.
Equilibrium characterization of the CEGG is left open.

With $F_t = \sum_{(u,v) \in E_1} f_t^{(u,v)}$ (total adoption), the per-step rewards are $R_1^{(t)} = \kappa F_t - c_1 \mathbf{1}[a_t^1 \neq \emptyset]$ and $R_2^{(t)} = p_H F_t - c_2 \mathbf{1}[a_t^2 \neq \emptyset]$, where $\kappa, p_H$ are per-unit benefits and $c_1, c_2$ are per-node installation costs.
Both rewards depend on $F_t$, which counts only operational adoption, requiring coordinated placements from both agents to render coupled positions operational.
The instantiation is common-interest rather than competitive: both rewards increase in the same adoption, differing only in scale and private cost.
A competing variant, two demand-side operators sharing a supply network, is a natural CEGG instance we leave open.

\subsection{Cross-Graph Attention}
\label{sec:algorithm}

Since Agent~1's optimal action at node $v$ depends on whether Agent~2 has activated nodes in $\Phi(v)$, we propose CGA: a $\Phi$-masked attention mechanism that scales with $|\Phi(v)|$ rather than $|V_2|$.

Each agent encodes its own network with a 3-layer GraphSAGE~\citep{hamilton2017inductive}, producing a hidden-dimension embedding $h_v$ for each node $v$ from input features $[m_v, \deg(v)/N, v/N]$ (installation status, normalized degree, node index). A linear advantage head $W_\psi$ produces per-node scores $\tilde\psi(s, v) = W_\psi h_v$; a linear value head on the mean-pooled embedding produces the scalar baseline $\hat V_t$, used to compute the empirical advantage $\hat A_t = \hat G_t - \hat V_t$ from the Monte Carlo (MC) return $\hat G_t = \sum_{s\geq t}\gamma^{s-t} r_s$ for variance-reduced policy gradient.

Let $h_v^{(1)}$ and $h_u^{(2)}$ denote the GraphSAGE embeddings of demand node $v \in V_1$ and supply node $u \in V_2$. For each demand node $v$, CGA computes a cross-graph context by scaled dot-product attention~\citep{vaswani2017attention} restricted to $v$'s coupling partners $\Phi(v)$:
\[
c_v = \sum_{u \in \Phi(v)} \alpha_{vu}\, W_V h_u^{(2)}, \qquad
\alpha_{vu} = \frac{\exp(q_v^\top k_u / \sqrt{d})}{\sum_{u' \in \Phi(v)} \exp(q_v^\top k_{u'} / \sqrt{d})},
\]
with queries $q_v = W_Q h_v^{(1)}$, keys $k_u = W_K h_u^{(2)}$, values $W_V h_u^{(2)}$, and $d$ the per-head key dimension ($d{=}16$: four heads over a $64$-dimensional embedding).
Restricting the softmax normalizer to $\Phi(v)$ structurally assigns $\alpha_{vu}=0$ for non-coupled supply nodes.
The context $c_v$ is concatenated with $h_v^{(1)}$ and projected to produce the coupling-aware embedding $\tilde h_v^{(1)} = \mathrm{ReLU}(W_O[h_v^{(1)} \| c_v])$, which replaces $h_v^{(1)}$ as input to the advantage and value heads.
Agent~2's embedding is computed symmetrically, attending over $\Phi^{-1}(u) = \{v \in V_1 : u \in \Phi(v)\}$.

\subsection{Training CGA: PPO and PMD Variants}
\label{sec:training}

CGA is an architectural component compatible with any policy-gradient method; we instantiate it with PPO~\citep{schulman2017proximal} (giving CGA-PPO) and with a deep-RL realization of policy mirror descent (PMD;~\citealp{lan2023policy}) in the spirit of MDPO~\citep{tomar2020mirror}, giving CGA-PMD.
We write $\psi$ for the idealized per-node advantage, $\tilde\psi_\theta$ for the parametric form of the neural advantage estimator from Section~\ref{sec:algorithm} ($\theta$ collects all CGA parameters), and $\hat\psi = \psi + \xi$ for its noisy realization at training time.
The pairing of CGA with PMD is not incidental: CGA's cross-graph signal makes Agent~$i$'s advantage estimate depend on Agent~$j$'s non-stationary embeddings, introducing a cross-graph noise term $\sigma_\Phi^2$ alongside the usual environment noise $\sigma_g^2$.
PMD's Kullback-Leibler (KL) regularization caps per-iteration drift under this noise; the clip update carries no such cap on the iterations where it does not bind (Section~\ref{sec:theory}), and Section~\ref{sec:experiments} reports the CGA-PPO breakdown observed at large $N$ under shared hyperparameters.

The standard PMD step minimizes a KL-regularized linearization of the policy objective:
\[
\pi_{k+1}^i = \arg\min_\pi -\langle \psi_i(s,\cdot),\, \pi(\cdot|s)\rangle + \lambda_k \,\mathrm{KL}(\pi\,\|\,\pi_k^i),
\]
with closed form $\pi_{k+1}^i(a|s) \propto \pi_k^i(a|s)\exp(\psi_i(s,a)/\lambda_k)$, where the regularizer $\lambda_k$ controls the per-step drift. Exact recovery requires tabular or linear parameterizations; with neural policies, the update must be approximated.

The CGA advantage head $\tilde\psi_{\theta^i}$ plays the role of $\psi_i$ above, and we sample actions from
\[
\pi_k^i(a|s) \;\propto\; \exp\bigl(\eta_k\,\tilde\psi_{\theta_k^i}(s,a)\bigr), \qquad \eta_k \;=\; \frac{\beta_k}{\lambda_k} \;=\; \frac{1}{\lambda\sqrt{k+1}}.
\]
This is the dual-averaging form of the mirror-descent step~\citep{ju2024policy}: a softmax of one accumulating score network rescaled each iteration, rather than the multiplicative $\pi_{k+1} \propto \pi_k\exp(\hat\psi_k/\lambda_k)$, which would require representing $\pi_k$ at every visited state, impractical on a combinatorial state space with a step-dependent action set.
The decreasing $\eta_k$ controls the effective logit magnitude $\eta_k\!\cdot\!\tilde\psi_\theta$: as $\tilde\psi_\theta$ grows via stochastic gradient descent, $\eta_k$ shrinks correspondingly, preventing premature commitment that would lock in cold-start patterns on the combinatorial action space. \citet{lan2023policy}'s linear-rate convergence is a single-agent, tabular result and does not directly transfer to our neural multi-agent setting; instead, the guarantee we state is the cumulative-drift bound of Theorem~\ref{thm:neural}, proved for the idealized update that this step approximates. The training loop is given in Algorithm~\ref{alg:cga}.

\begin{algorithm}[h]
\caption{CGA-PMD: Deep-RL PMD training loop for CGA. Loss terms, hyperparameters and implementation notes in App.~\ref{app:experimental}.}
\label{alg:cga}
\begin{algorithmic}[1]
\REQUIRE CEGG instance; iterations $K$; schedule $\beta_k{=}k{+}1$, $\lambda_k{=}\lambda(k{+}1)^{3/2}$, $\eta_k{=}1/(\lambda\sqrt{k{+}1})$.
\STATE Initialize CGA advantage network $\tilde\psi_{\theta^i}$ for each agent $i$.
\FOR{$k = 0, 1, \ldots, K{-}1$}
    \STATE Sample joint rollout: $a_t^i \sim \pi_k^i(\cdot|s_t) \propto \exp(\eta_k \tilde\psi_{\theta_k^i}(s_t,\cdot))$ for each agent $i$; collect $\{(s_t, a_t^1, a_t^2, r_t)\}$ with shared reward $r_t$.
    \STATE Compute MC return $\hat G_t = \sum_{s\geq t}\gamma^{s-t} r_s$, baseline $\hat V_t$ (detached value-head output), and per-episode-normalized advantage $\hat A_t = \hat G_t - \hat V_t$.
    \STATE Take one Adam step on the joint objective over $\theta = (\theta^1,\theta^2)$:
    \[
    \mathcal L(\theta) \;=\; -\,\tfrac{w_k}{T}\!\sum_t \bigl[\log\pi_k^1(a_t^1|s_t) + \log\pi_k^2(a_t^2|s_t)\bigr]\hat A_t \;+\; \tfrac{1}{2T}\!\sum_t \bigl(\bar\psi_{\theta^1}(s_t)+\bar\psi_{\theta^2}(s_t)-\hat G_t\bigr)^2,
    \] where $\bar\psi_{\theta^i}(s) := |V_i|^{-1}\sum_{v\in V_i}\tilde\psi_{\theta^i}(s,v)$ and $w_k = \beta_k/\!\sum_{j\le k}\!\beta_j$.
\ENDFOR
\RETURN $\pi_K^1, \pi_K^2$
\end{algorithmic}
\end{algorithm}

\begin{table}[h]
\centering
\caption{Methods compared, by observation setting and optimizer.}
\label{tab:methods}
\small
\begin{tabular*}{\textwidth}{@{\extracolsep{\fill}}lllc@{}}
\toprule
\textbf{Method} & \textbf{Observation} & \textbf{Optimizer} & \textbf{Reference} \\
\midrule
IPPO            & own only           & PPO  & \citep{dewitt2020independent} \\
Indep-PMD       & own only           & PMD  &  \\
CGA-PPO         & own $+$ masked     & PPO  &  \\
CGA-PMD         & own $+$ masked     & PMD  & this work \\
MAPPO-Het       & own $+$ unmasked   & PPO  & \citep{mei2026multi} \\
DGN-PMD         & own $+$ unmasked   & PMD  &  \\
\bottomrule
\end{tabular*}
\end{table}

\section{Experiments}
\label{sec:experiments}

We instantiate the CEGG across three reward calibrations: hydrogen freight (Agent~1 places hydrogen refueling stations on a freight logistics network $G_1$, Agent~2 places hydrogen production plants on $G_2$; reward via fuel-cell-truck Bass diffusion), EV-grid (Agent~1 places direct current fast chargers on a road network $G_1$, Agent~2 places grid substations on $G_2$; reward via EV-adoption Bass diffusion), and vaccination (Agent~1 places clinics on a demand network $G_1$, Agent~2 places cold-chain depots on $G_2$; reward via vaccine-uptake Bass diffusion).
These instantiations are stylized: the Bass form, its parameters and the costs are literature estimates, not operator data, and deployment would need calibration against an operator's records (App.~\ref{app:cr_pq} varies them).

For network instances we use open-source Transportation Network Test Problems (TNTP) ~\citep{stabler2024transportation}, a curated repository of real city road networks in which nodes are intersections and edges are road segments.
We use breadth-first search (BFS) subsamples of ChicagoSketch, nested subgraphs grown by breadth-first expansion outward from the highest-degree node, at $N{\in}\{50,100,150,200,250,300\}$ for controlled scaling, and five real city networks spanning $17{\times}$ in size (Sioux Falls $24$ to BerlinMitte $397$ nodes).
Supply graphs use top-degree nodes (i.e., $N_2{=}\lceil N_1/3 \rceil$, with coupling radius $d_{\max}{=}2$ --- $\Phi(v) = \{u \in V_2 : d_{G_1}(v, u) \leq 2\}$, the supply nodes within graph-hop distance 2 of demand $v$).

The six methods of Table~\ref{tab:methods} are evaluated on every cell with 10 seeds per (method, network) cell.
The comparison is a factorial over architecture and optimizer rather than a survey: HAPPO~\citep{kuba2022happo}, MADDPG~\citep{lowe2017multi}, and QMIX~\citep{rashid2020monotonic} assume a shared observation or a joint action space over one graph, so they do not apply to two separately expanded graphs.
We use patience-based early stopping (patience 100, max 1000 episodes); the full network list, hyperparameters and the significance protocol are in App.~\ref{app:experimental}.

A seed counts as a ``success'' if the trained policy's evaluation reward exceeds a productive threshold ($5{\cdot}10^{7}$ at hydrogen calibration; scaled by $\kappa$ for EV-grid and vaccination).
Sub-threshold seeds are dominated by a cold-start collapse in which the policy hits a no-learning floor (typically $-4.8{\cdot}10^{7}$: installation cost without adoption) and never escapes; this is failure of the trained policy, not training-time non-convergence (early stopping does fire).
``$k/n$'' notation in figures and tables denotes $k$ successful seeds out of $n$.
No seed lands between $4.54{\cdot}10^{7}$ and $8.46{\cdot}10^{7}$, so the counts do not depend on where the line is drawn, and reward gaps are significant in $23$ of $24$ cells by paired Wilcoxon (App.~\ref{app:cr_stats}).
The experiments are organized along three borderlines: scaling at fixed coupling structure (Section~\ref{sec:scaling}), coupling structure at fixed scale (Section~\ref{sec:coupling}), and method-by-method behavior on real city networks across all three reward domains (Section~\ref{sec:realnets}).

\subsection{Scaling Borderline}
\label{sec:scaling}

\begin{figure}[h]
\centering
\includegraphics[width=\textwidth]{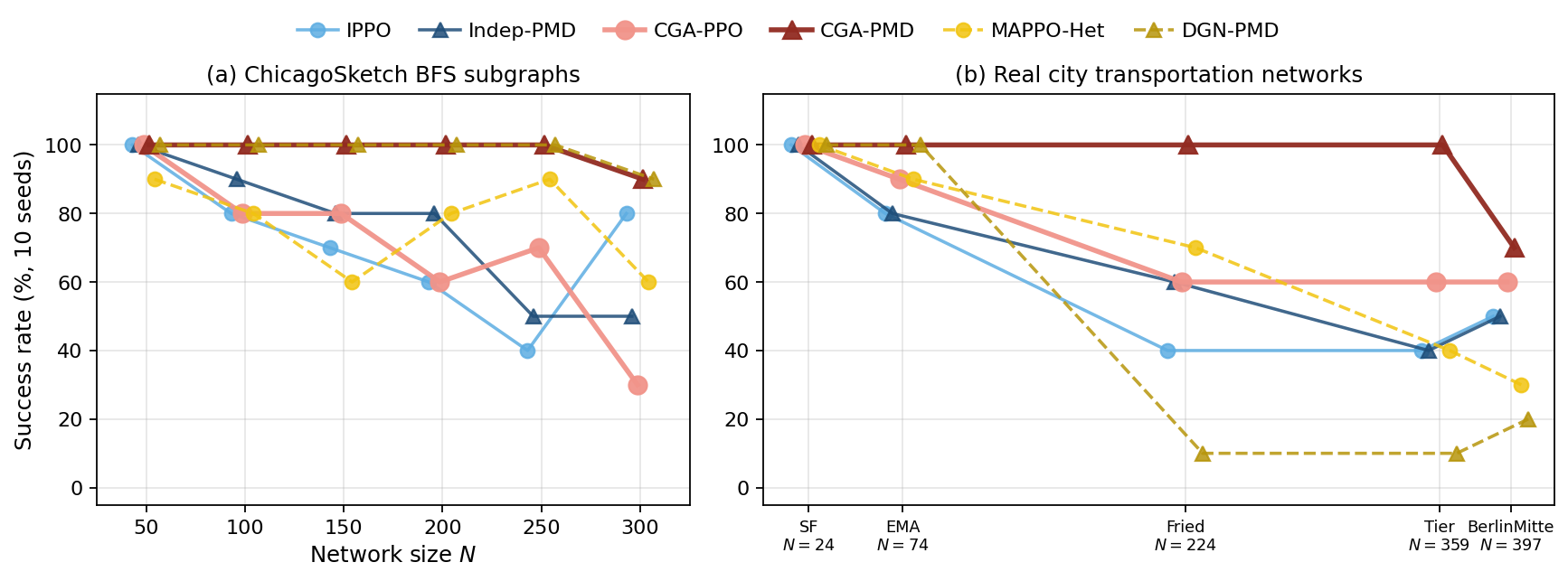}
\caption{Scaling borderline (10 seeds): (a) ChicagoSketch BFS subgraphs; (b) real city transportation networks.}
\label{fig:scaling}
\end{figure}

\begin{figure}[h]
\centering
\includegraphics[width=0.80\textwidth]{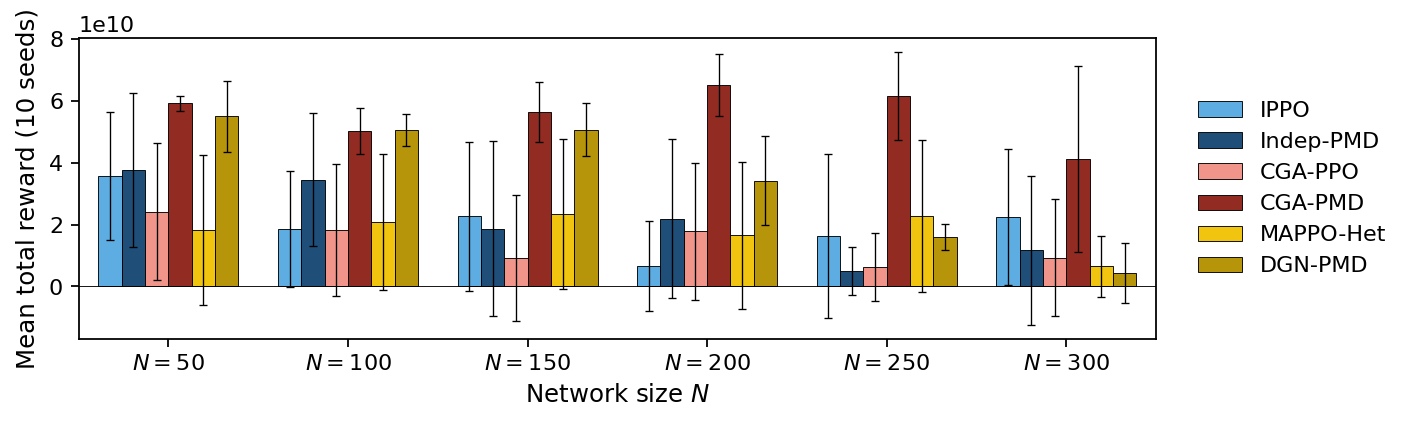}
\caption{Mean total reward by network size on ChicagoSketch BFS subgraphs ($d_{\max}{=}2$, 10 seeds; error bars: std). CGA-PMD has the highest mean from $N{=}150$ upward; at $N{\leq}100$ the leading methods are within seed noise of one another.}
\label{fig:scaling_meanstd}
\end{figure}

At a fixed coupling cardinality ($d_{\max}{=}2$), we sweep action-space scale on controlled BFS subgraphs of ChicagoSketch ($N{\in}\{50,100,150,200,250,300\}$) and complement this with five real city transportation networks ($N{\in}\{24,74,224,359,397\}$) at hydrogen instantiation.
Figure~\ref{fig:scaling} shows success rate per method per scale on both axes; mean reward and seed-level variability are summarized in Figure~\ref{fig:scaling_meanstd}.

Reading both panels of Figure~\ref{fig:scaling} alongside the mean-reward sweep of Figure~\ref{fig:scaling_meanstd} exposes three regimes.
Small (BFS $N{\leq}100$; Sioux Falls and EMA on real): all six methods reach $\geq 80\%$ success and CGA-PMD's mean reward is on par with the next-best baseline.
Intermediate (BFS $N{=}150$--$200$; Friedrichshain $N{=}224$ on real): CGA-PMD stays at $100\%$ while IPPO, Indep-PMD, and CGA-PPO drop into the $40$--$80\%$ band, and unmasked DGN-PMD already collapses to $10\%$ on Friedrichshain.
Above-boundary (BFS $N{\geq}250$; Tiergarten and BerlinMitte on real): the CGA-PMD vs.\ CGA-PPO gap opens to $90\%$ vs.\ $30\%$ at BFS $N{=}300$, and on real networks $\Phi$-masked CGA-PMD retains $100\%$ at Tiergarten ($N{=}359$) and $70\%$ at BerlinMitte ($N{=}397$) while unmasked methods (DGN-PMD, MAPPO-Het) drop to $\leq 40\%$, the $\Phi$-mask becomes the decisive structural feature.

\subsection{Coupling Borderline}
\label{sec:coupling}

\begin{figure}[h]
    \centering
    \includegraphics[width=\textwidth]{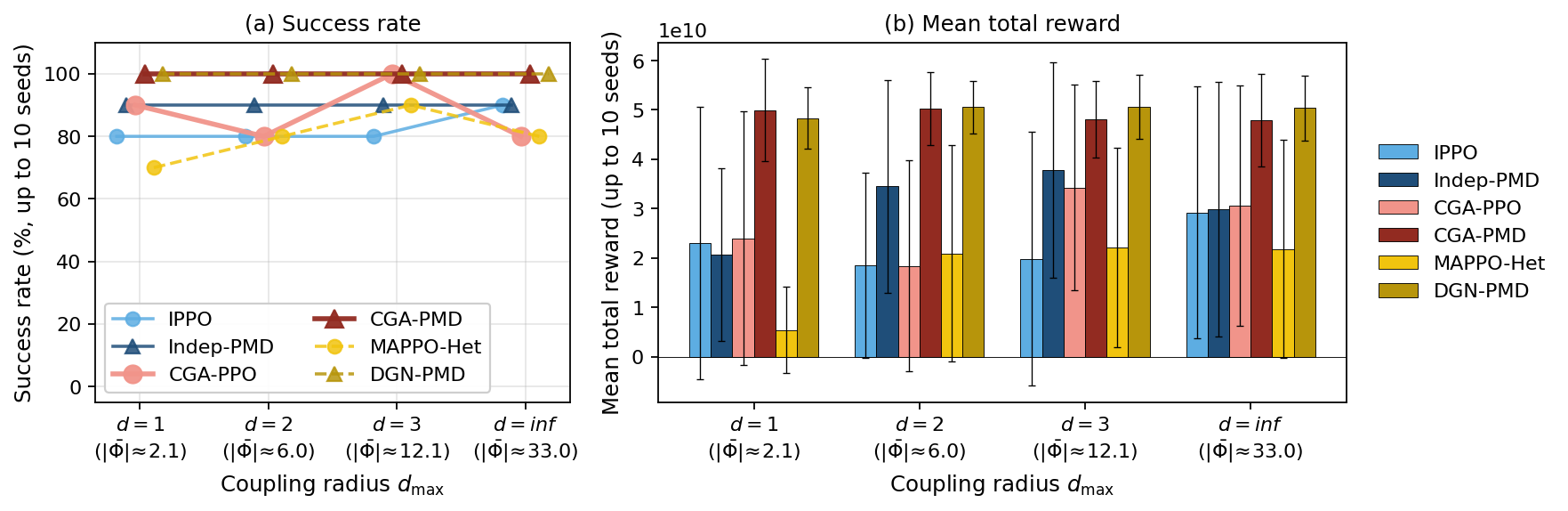}
    \caption{Coupling borderline at $N{=}100$ (ChicagoSketch BFS, 10 seeds): (a) success rate and (b) mean total reward (error bars: std) vs.\ coupling radius $d_{\max}$ for the six methods of Table~\ref{tab:methods}.}
    \label{fig:coupling}
\end{figure}

At a fixed scale (ChicagoSketch BFS at $N{=}100$), we vary the coupling radius $d_{\max}{\in}\{1,2,3,\infty\}$ to sweep average coupling cardinality $\bar{|\Phi|}{\in}\{2.1,6.0,12.1,33.0\}$.
Figure~\ref{fig:coupling} shows success rate per method per density.
The same six primary methods are evaluated.

At fixed $N{=}100$, every method stays at or above $70\%$ success at every $d_{\max}$ and CGA-PMD and DGN-PMD reach $100\%$ at all four, so the success-rate axis does not separate the methods (Figure~\ref{fig:coupling}a).
The mean-reward axis does (Figure~\ref{fig:coupling}b): CGA-PMD and DGN-PMD reach ${\sim}5{\cdot}10^{10}$ at every $d_{\max}$ while CGA-PPO, IPPO, Indep-PMD, and MAPPO-Het cluster between $0.5$ and $3.8{\cdot}10^{10}$, a ${\sim}2{\times}$ gap on average that is independent of coupling density.
The gap is not the optimizer alone: Indep-PMD, PMD without cross-graph attention, sits in the lower cluster, so what separates the leading pair is the pairing.
Within that pair the $\Phi$ mask does not separate the two at this scale; its effect appears with scale rather than sparsity, which is what the bias term of Section~\ref{sec:theory} predicts and what $1/|\Phi|$ variance scaling alone cannot, since that favors attending more widely.
What this sweep establishes is that at $N{=}100$ the coupling radius does not change which methods are viable and that CGA-PMD is one of the leading viable methods.

\subsection{Reward-Domain Robustness}
\label{sec:realnets}

To verify that the regime structure of Section~\ref{sec:scaling} is not specific to a single reward calibration, we instantiate the CEGG under three: hydrogen freight ($\kappa{=}\$8{,}500$/truck CO$_2$ saving), EV-grid ($\kappa{=}\$1{,}275$/EV/yr), and vaccination ($\kappa{=}\$500$/dose).
Bass-diffusion and cost parameters are calibrated from the literature~\citep{becker2009ev, bernal2024dcfc, epa2023scghg, gorbachuk2022vaccine, kohli2021covid, leidner2019adult, massiani2015bass, smith2015evse, yarnoff2022clinics}; full per-domain breakdowns in App.~\ref{app:reward_params}.

\begin{figure}[t]
\centering
\includegraphics[width=0.75\textwidth]{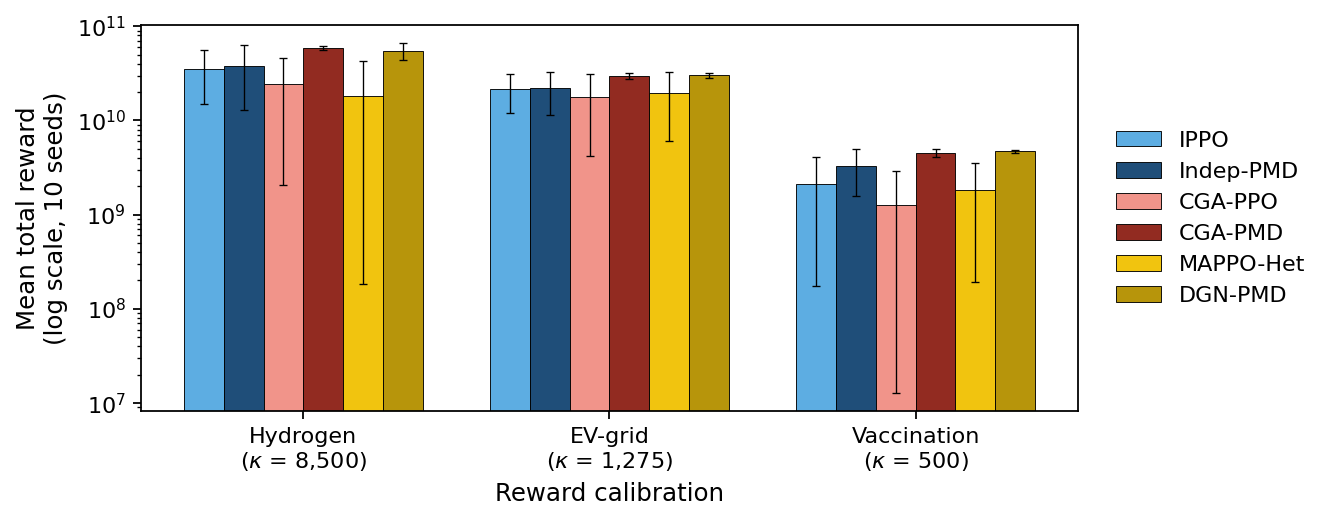}
\caption{Reward-calibration robustness at the $N{=}50$ control baseline (ChicagoSketch BFS, 10 seeds; mean total reward (error bars: std), log scale). The method ordering is preserved across $\kappa{\in}\{\$500,\$1{,}275,\$8{,}500\}$ (a $17{\times}$ magnitude range): the two attention-based PMD methods, CGA-PMD and DGN-PMD, lead in all three domains by $1.3$--$1.6\times$ over the best other method, isolating the architectural advantage from any reward-magnitude effect.}
\label{fig:domains}
\end{figure}

Figure~\ref{fig:domains} shows mean total reward on ChicagoSketch BFS at $N{=}50$, chosen as the control baseline because it is the small-regime point where all six methods reach productive learning (no method collapses to the cold-start floor), so any ordering on this slice reflects architectural differences rather than confounds from differential failure rates.
The ordering is preserved across the $17{\times}$ magnitude range of $\kappa$: the two attention-based PMD methods lead in all three domains, by $1.3$--$1.6\times$ over the best other method, isolating the architectural advantage from any reward-magnitude effect.

\section{Theoretical Analysis: Why CGA Pairs with PMD}
\label{sec:theory}

CGA injects cross-graph signal at the cost of a cross-graph noise term $\sigma_\Phi^2$ alongside standard environment noise $\sigma_g^2$.
We show that $\sigma_\Phi^2$ is architecturally bounded (Lemma~\ref{lem:cross_var}) and that PMD's per-step KL regularization caps its cumulative effect uniformly in iterations $K$ (Lemma~\ref{lem:single_step}, Theorem~\ref{thm:neural}); for the clip update we prove the complementary negative statement on the iterations where the clip does not bind (Proposition~\ref{prop:ppo_drift}).
Proofs are in App.~\ref{app:proof}.

\begin{lemma}[Cross-graph noise scaling]
\label{lem:cross_var}
For a demand node $v \in V_1$ with coupling partners $\Phi(v) \subseteq V_2$ (the set of supply nodes structurally coupled to $v$), the cross-graph context vector is
\[
c_v \;=\; \sum_{u \in \Phi(v)} \alpha_{vu}\; W_V\; h_u^{(2)},
\]
where $h_u^{(2)} \in \mathbb{R}^d$ is the GraphSAGE embedding of supply node $u$ (hidden embedding dimension $d$), $W_V \in \mathbb{R}^{d \times d}$ is the attention value-projection matrix with squared Frobenius norm $\|W_V\|_F^2 := \sum_{i,j} (W_V)_{ij}^2$, and $\alpha_{vu} \geq 0$ is the attention weight from $v$ to $u$ with $\sum_{u \in \Phi(v)} \alpha_{vu} = 1$.
The cross-graph noise variance is bounded by
\[
\sigma_\Phi^2 \;:=\; \mathbb{E}\|c_v\|^2 \;\leq\; \frac{\sigma_h^2\, \|W_V\|_F^2}{|\Phi(v)|},
\]
with \emph{equality} at uniform attention $\alpha_{vu} = 1/|\Phi(v)|$, under mean-zero supply embeddings with isotropic variance $\mathrm{Cov}(h_u^{(2)}) = \sigma_h^2 I$ for some scalar $\sigma_h^2 > 0$.
\end{lemma}

\emph{Proof sketch.} Expand $\mathbb{E}\|c_v\|^2$; pairwise-uncorrelated supply embeddings kill the cross-terms ($u \neq u'$), leaving $\sigma_h^2 \|W_V\|_F^2 \sum_u \alpha_{vu}^2$, which is closed at uniform attention by Cauchy--Schwarz on the simplex. \hfill$\square$

Lemma~\ref{lem:cross_var} bounds the variance term only, and read alone it favors attending more widely.
Allowing partner and non-partner embeddings to have different means adds a bias term: with $\omega$ the attention mass an unmasked model places off $\Phi(v)$ and $\delta$ the separation between the two group means, the context error is $\omega^2\delta^2$ plus the variance of Lemma~\ref{lem:cross_var}.
Masking sets $\omega = 0$, so it helps exactly when $\delta^2 > \sigma_h^2\|W_V\|_F^2/(|\Phi|(1-\rho))$, $\rho = |\Phi|/|V_2|$.

The optimizer side admits a complementary per-step bound depending only on the sup-norm of the advantage estimate, not its variance:

\begin{lemma}[Single-step KL drift for PMD]
\label{lem:single_step}
Let $\hat\psi_k$ be the noisy advantage estimate at iteration $k$, and let $B > 0$ be any constant satisfying $\|\hat\psi_k\|_\infty := \sup_{s,a}|\hat\psi_k(s,a)| \leq B$.
Let $\lambda_k > 0$ be the PMD regularizer.
Then the PMD update $\pi_{k+1}^i(a|s) \propto \pi_k^i(a|s)\exp(\hat\psi_k/\lambda_k)$ satisfies
\[
\mathrm{KL}\bigl(\pi_{k+1}^i\,\|\,\pi_k^i\bigr) \;\leq\; \frac{B^2}{2\lambda_k^2} + O\bigl(B^3/\lambda_k^3\bigr),
\]
independent of advantage variance.
\end{lemma}

\emph{Proof sketch.} The KL identity $\mathrm{KL}(\pi_{k+1}\|\pi_k) = \mathbb{E}_{\pi_{k+1}}[g] - \log Z$ with $g := \hat\psi_k/\lambda_k$, combined with Hoeffding's lemma controlling $\log Z - \mathbb{E}_\pi[g] \leq B^2/(2\lambda_k^2)$ and Bregman-duality symmetrization, yields the claim. \hfill$\square$

Combining Lemma~\ref{lem:cross_var} (cross-graph noise) and Lemma~\ref{lem:single_step} (per-step KL drift) with an $L$-Lipschitz GraphSAGE backbone yields the main result:

\begin{theorem}[Architectural drift bound for the idealized CGA-PMD update]
\label{thm:neural}
Under Lemma~\ref{lem:cross_var}'s assumptions and an $L$-Lipschitz GraphSAGE backbone (i.e., $\|f(x){-}f(x')\| \leq L\|x{-}x'\|$ for all $x, x'$, with $f$ the encoder and $L > 0$ a constant), the neural CGA advantage estimator $\hat\psi^\theta(s,v) = f_\theta(h_v^{(1)}, c_v)$ admits the architectural sup-norm decomposition
\[
\|\hat\psi^\theta\|_\infty \;\leq\; \tilde B \;:=\; B_{\mathrm{MDP}} \,+\, \epsilon_{\mathrm{local}} \,+\, \frac{L\,\sigma_h\,\|W_V\|_F}{\sqrt{|\Phi|}},
\]
where $B_{\mathrm{MDP}}{=}R/(1{-}\gamma)$ ($R$ the per-step reward bound, $\gamma$ the discount factor) and $\epsilon_{\mathrm{local}}$ is the GraphSAGE approximation error.
Consequently, under the schedule $\lambda_k = \lambda(k{+}1)^{3/2}$, the idealized multiplicative update $\pi^\theta_{k+1}(a|s) \propto \pi^\theta_k(a|s)\exp(\hat\psi^\theta_k/\lambda_k)$ obeys
\[
\sum_{k=0}^{K-1}\mathrm{KL}\bigl(\pi^\theta_{k+1}\|\pi^\theta_k\bigr) \;\leq\; \frac{\tilde B^2 \zeta(3)}{2\lambda^2}+O\bigl(\tilde B^3/\lambda^3\bigr), \qquad \text{uniformly in } K,
\]
where $\zeta(3):=\sum_{n\geq 1} n^{-3}\approx 1.202$; Cauchy-in-TV iterate convergence follows (Corollary~\ref{cor:cauchy}).
\end{theorem}

\emph{Proof sketch.} Lemma~\ref{lem:cross_var}'s noise bound propagates through the $L$-Lipschitz GraphSAGE backbone into the sup-norm decomposition $\tilde B$.
Combining Lemma~\ref{lem:single_step}'s per-step bound with $\sum_{k\geq 0} 1/(k{+}1)^3 = \zeta(3)$ under $\lambda_k = \lambda(k{+}1)^{3/2}$ gives the $K$-uniform ceiling.
The third decomposition term $L\sigma_h\|W_V\|_F/\sqrt{|\Phi|}$ ties the bound to coupling cardinality. \hfill$\square$

\begin{figure}[t]
\centering
\begin{subfigure}[b]{0.467\textwidth}
    \includegraphics[width=\linewidth]{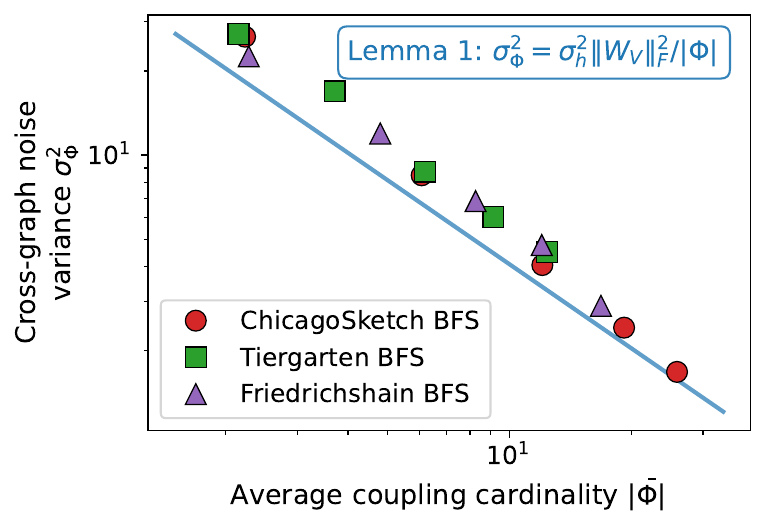}
    \caption{}
    \label{fig:lemma1}
\end{subfigure}
\hfill
\begin{subfigure}[b]{0.515\textwidth}
    \includegraphics[width=\linewidth]{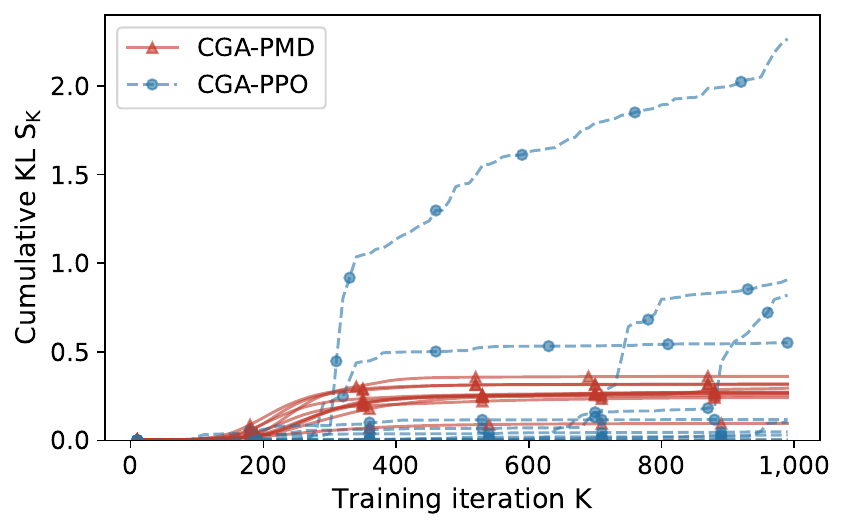}
    \caption{}
    \label{fig:stationarity}
\end{subfigure}
\caption{Drift validation. (a) Cross-graph noise variance vs.\ coupling cardinality; the Lemma~\ref{lem:cross_var} prediction matches across three graph types ($R^2{>}0.99$). (b) Cumulative KL $S_K$ on ChicagoSketch BFS $N{=}100$ (10 seeds, $K$ to $1{,}000$): CGA-PMD flattens as $K$ grows, consistent with Theorem~\ref{thm:neural}, while CGA-PPO shows upward drift with wide seed-dependent variability.}
\label{fig:drift_validation}
\end{figure}

The theorem analyzes the idealized multiplicative update under $\lambda_k = \lambda(k{+}1)^{3/2}$, whereas Algorithm~\ref{alg:cga} implements that step in dual-averaging form, with $\eta_k = \beta_k/\lambda_k$ and no explicit KL anchor to $\pi_k$.
The two belong to the same mirror-descent family and share $\lambda_k$, but the bound is stated for the idealized update; establishing it for the dual-averaging form is left open.
Figure~\ref{fig:stationarity} reports the drift of the implemented algorithm.

Lemma~\ref{lem:single_step} specializes the broader Bregman-divergence framework of~\citet{lan2023policy} to the KL case (the natural choice given CGA's softmax attention); the general analysis extends to other distance-generating functions.
Lemma~\ref{lem:cross_var} and the architectural decomposition in Theorem~\ref{thm:neural} are CGA-specific.
Three empirical observations corroborate the analysis.
First, Lemma~\ref{lem:cross_var}'s $1/|\Phi|$ scaling holds across three graph types with $R^2{>}0.99$ (Figure~\ref{fig:lemma1}).
Second, CGA-PMD's cumulative KL flattens as $K$ grows while CGA-PPO's grows with wide seed-dependent variability, consistent with Theorem~\ref{thm:neural} and Proposition~\ref{prop:ppo_drift} respectively (Figure~\ref{fig:stationarity}).
Third, the predicted gap manifests at scale: CGA-PMD attains $90\%$ success vs.\ CGA-PPO's $30\%$ at $N{=}300$, and on the irregular real networks CGA-PMD attains the highest success count on all five (Figure~\ref{fig:scaling}; Sections~\ref{sec:scaling} and~\ref{sec:realnets}).

\section{Conclusion and Discussion}
\label{sec:conclusion}

The Co-Evolving Graph Game (CEGG) captures sequential co-design over two coupled graphs, in a setting not addressed by facility location, interdependent-network analysis or graph-MARL individually.
We propose CGA-PMD: $\Phi$-masked cross-graph attention paired with a deep-RL realization of policy mirror descent.
The two are a pairing: the attention introduces the partner-graph noise the KL regularization controls and the grid ablates it.

CGA-PMD keeps a $10/10$ success rate through $N{=}250$ and $9/10$ at $N{=}300$, and attains the highest success count on each of the five real city networks, where the unmasked DGN-PMD falls to $1/10$ on two. On the BFS sweep it also leads on mean reward from $N{=}150$ upward, by $1.1$--$1.9\times$ once collapsed seeds are removed from every method.
These results establish that CGA-PMD is scalable and reliable at large action-space scale.

There are limitations to this study that naturally open avenues for further work.
Theorem~\ref{thm:neural} is proved for the idealized multiplicative update, and Lemma~\ref{lem:single_step} bounds single-step drift rather than a convergence rate.
Proposition~\ref{prop:ppo_drift} is scoped to the iterations on which the clip does not bind.
Extending the bound to the implemented update is the natural next theoretical step.
None of these qualifications affects the comparison the experiments support: as scale grows or coupling becomes irregular, CGA-PMD is the method that stays reliable across three reward domains and five real city networks.
The CEGG formulation and the evaluation protocol introduced here give that continuation a setting to build on.

\begin{ack}
This work was supported by the Net Zero Freight Systems program at the Georgia Tech Supply Chain and Logistics Institute, and by the Office of Naval Research under grant N000142412654.

The authors declare no competing interests.
\end{ack}

\bibliographystyle{plainnat}

\newpage
\appendix

\section{Comprehensive Experimental Results}
\label{app:results}

Per-cell numbers behind the figures of Section~\ref{sec:experiments}: the controlled scaling sweep (Table~\ref{tab:appA_bfs}), the coupling-density sweep of Figure~\ref{fig:coupling} (Table~\ref{tab:appA_density}), the five real city networks of Figure~\ref{fig:scaling}b (Table~\ref{tab:appA_real}), the three reward calibrations of Figure~\ref{fig:domains} (Table~\ref{tab:appA_3domain}), the Lemma~\ref{lem:cross_var} validation of Figure~\ref{fig:lemma1} (Table~\ref{tab:appA_lemma1}), and the cumulative-KL trajectories of Figure~\ref{fig:stationarity} (Table~\ref{tab:appA_kl}). All cells are 10 seeds unless stated otherwise.

\begin{table}[H]
\centering
\scriptsize
\setlength{\tabcolsep}{3pt}
\caption{Full results: ChicagoSketch BFS scaling (hydrogen calibration, $d_{\max}{=}2$). Cells show mean$\pm$std and (success$>5{\cdot}10^{7}$ / total seeds).}
\label{tab:appA_bfs}
\resizebox{\textwidth}{!}{%
\begin{tabular}{@{}lcccccc@{}}
\toprule
\textbf{Method} & $N{=}50$ & $N{=}100$ & $N{=}150$ & $N{=}200$ & $N{=}250$ & $N{=}300$ \\
\midrule
IPPO & $3.56{\cdot}10^{10}{\pm}2.1{\cdot}10^{10}$ (10/10) & $1.86{\cdot}10^{10}{\pm}1.9{\cdot}10^{10}$ (8/10) & $2.27{\cdot}10^{10}{\pm}2.4{\cdot}10^{10}$ (7/10) & $6.61{\cdot}10^{9}{\pm}1.5{\cdot}10^{10}$ (6/10) & $1.63{\cdot}10^{10}{\pm}2.6{\cdot}10^{10}$ (4/10) & $2.25{\cdot}10^{10}{\pm}2.2{\cdot}10^{10}$ (8/10) \\
Indep-PMD & $3.77{\cdot}10^{10}{\pm}2.5{\cdot}10^{10}$ (10/10) & $3.45{\cdot}10^{10}{\pm}2.1{\cdot}10^{10}$ (9/10) & $1.87{\cdot}10^{10}{\pm}2.8{\cdot}10^{10}$ (8/10) & $2.18{\cdot}10^{10}{\pm}2.6{\cdot}10^{10}$ (8/10) & $5.03{\cdot}10^{9}{\pm}7.9{\cdot}10^{9}$ (5/10) & $1.17{\cdot}10^{10}{\pm}2.4{\cdot}10^{10}$ (5/10) \\
CGA-PPO & $2.42{\cdot}10^{10}{\pm}2.2{\cdot}10^{10}$ (10/10) & $1.84{\cdot}10^{10}{\pm}2.1{\cdot}10^{10}$ (8/10) & $9.19{\cdot}10^{9}{\pm}2.0{\cdot}10^{10}$ (8/10) & $1.78{\cdot}10^{10}{\pm}2.2{\cdot}10^{10}$ (6/10) & $6.32{\cdot}10^{9}{\pm}1.1{\cdot}10^{10}$ (7/10) & $9.32{\cdot}10^{9}{\pm}1.9{\cdot}10^{10}$ (3/10) \\
CGA-PMD & $5.92{\cdot}10^{10}{\pm}2.5{\cdot}10^{9}$ (10/10) & $5.02{\cdot}10^{10}{\pm}7.4{\cdot}10^{9}$ (10/10) & $5.63{\cdot}10^{10}{\pm}9.6{\cdot}10^{9}$ (10/10) & $6.51{\cdot}10^{10}{\pm}1.0{\cdot}10^{10}$ (10/10) & $6.17{\cdot}10^{10}{\pm}1.4{\cdot}10^{10}$ (10/10) & $4.13{\cdot}10^{10}{\pm}3.0{\cdot}10^{10}$ (9/10) \\
MAPPO-Het & $1.83{\cdot}10^{10}{\pm}2.4{\cdot}10^{10}$ (9/10) & $2.09{\cdot}10^{10}{\pm}2.2{\cdot}10^{10}$ (8/10) & $2.35{\cdot}10^{10}{\pm}2.4{\cdot}10^{10}$ (6/10) & $1.65{\cdot}10^{10}{\pm}2.4{\cdot}10^{10}$ (8/10) & $2.27{\cdot}10^{10}{\pm}2.5{\cdot}10^{10}$ (9/10) & $6.56{\cdot}10^{9}{\pm}9.9{\cdot}10^{9}$ (6/10) \\
DGN-PMD & $5.50{\cdot}10^{10}{\pm}1.1{\cdot}10^{10}$ (10/10) & $5.06{\cdot}10^{10}{\pm}5.3{\cdot}10^{9}$ (10/10) & $5.06{\cdot}10^{10}{\pm}8.6{\cdot}10^{9}$ (10/10) & $3.42{\cdot}10^{10}{\pm}1.4{\cdot}10^{10}$ (10/10) & $1.59{\cdot}10^{10}{\pm}4.1{\cdot}10^{9}$ (10/10) & $4.45{\cdot}10^{9}{\pm}9.7{\cdot}10^{9}$ (9/10) \\
\bottomrule
\end{tabular}}
\end{table}
\begin{table}[H]
\centering
\scriptsize
\setlength{\tabcolsep}{3pt}
\caption{Full results: coupling density at $N{=}100$ (ChicagoSketch BFS). $\bar{|\Phi|}$ values: 2.1, 6.0, 12.1, 33.0 for $d_{\max}{\in}\{1,2,3,\infty\}$.}
\label{tab:appA_density}
\resizebox{\textwidth}{!}{%
\begin{tabular}{@{}lcccc@{}}
\toprule
\textbf{Method} & $d_{\max}{=}1$ & $d_{\max}{=}2$ & $d_{\max}{=}3$ & $d_{\max}{=}inf$ \\
\midrule
IPPO & $2.30{\cdot}10^{10}{\pm}2.8{\cdot}10^{10}$ (8/10) & $1.86{\cdot}10^{10}{\pm}1.9{\cdot}10^{10}$ (8/10) & $1.98{\cdot}10^{10}{\pm}2.6{\cdot}10^{10}$ (8/10) & $2.92{\cdot}10^{10}{\pm}2.6{\cdot}10^{10}$ (9/10) \\
Indep-PMD & $2.07{\cdot}10^{10}{\pm}1.8{\cdot}10^{10}$ (9/10) & $3.45{\cdot}10^{10}{\pm}2.1{\cdot}10^{10}$ (9/10) & $3.78{\cdot}10^{10}{\pm}2.2{\cdot}10^{10}$ (9/10) & $2.98{\cdot}10^{10}{\pm}2.6{\cdot}10^{10}$ (9/10) \\
CGA-PPO & $2.40{\cdot}10^{10}{\pm}2.6{\cdot}10^{10}$ (9/10) & $1.84{\cdot}10^{10}{\pm}2.1{\cdot}10^{10}$ (8/10) & $3.43{\cdot}10^{10}{\pm}2.1{\cdot}10^{10}$ (10/10) & $3.05{\cdot}10^{10}{\pm}2.4{\cdot}10^{10}$ (8/10) \\
CGA-PMD & $5.00{\cdot}10^{10}{\pm}1.0{\cdot}10^{10}$ (10/10) & $5.02{\cdot}10^{10}{\pm}7.4{\cdot}10^{9}$ (10/10) & $4.81{\cdot}10^{10}{\pm}7.7{\cdot}10^{9}$ (10/10) & $4.79{\cdot}10^{10}{\pm}9.4{\cdot}10^{9}$ (10/10) \\
MAPPO-Het & $5.41{\cdot}10^{9}{\pm}8.8{\cdot}10^{9}$ (7/10) & $2.09{\cdot}10^{10}{\pm}2.2{\cdot}10^{10}$ (8/10) & $2.21{\cdot}10^{10}{\pm}2.0{\cdot}10^{10}$ (9/10) & $2.18{\cdot}10^{10}{\pm}2.2{\cdot}10^{10}$ (8/10) \\
DGN-PMD & $4.83{\cdot}10^{10}{\pm}6.2{\cdot}10^{9}$ (10/10) & $5.06{\cdot}10^{10}{\pm}5.3{\cdot}10^{9}$ (10/10) & $5.06{\cdot}10^{10}{\pm}6.5{\cdot}10^{9}$ (10/10) & $5.04{\cdot}10^{10}{\pm}6.6{\cdot}10^{9}$ (10/10) \\
\bottomrule
\end{tabular}}
\end{table}
\begin{table}[H]
\centering
\scriptsize
\setlength{\tabcolsep}{3pt}
\caption{Full results: real city transportation networks (hydrogen calibration). Networks ordered by size $N$.}
\label{tab:appA_real}
\resizebox{\textwidth}{!}{%
\begin{tabular}{@{}lccccc@{}}
\toprule
\textbf{Method} & SF\,(24) & EMA\,(74) & Fried\,(224) & Tier\,(359) & BerlinMitte\,(397) \\
\midrule
IPPO & $3.88{\cdot}10^{10}{\pm}1.0{\cdot}10^{10}$ (10/10) & $8.71{\cdot}10^{9}{\pm}7.5{\cdot}10^{9}$ (8/10) & $3.60{\cdot}10^{11}{\pm}7.6{\cdot}10^{11}$ (4/10) & $1.76{\cdot}10^{11}{\pm}3.7{\cdot}10^{11}$ (4/10) & $1.08{\cdot}10^{11}{\pm}1.3{\cdot}10^{11}$ (5/10) \\
Indep-PMD & $3.85{\cdot}10^{10}{\pm}1.2{\cdot}10^{10}$ (10/10) & $6.55{\cdot}10^{9}{\pm}5.9{\cdot}10^{9}$ (8/10) & $1.42{\cdot}10^{11}{\pm}3.8{\cdot}10^{11}$ (6/10) & $8.97{\cdot}10^{9}{\pm}2.8{\cdot}10^{10}$ (4/10) & $7.04{\cdot}10^{11}{\pm}9.1{\cdot}10^{11}$ (5/10) \\
CGA-PPO & $3.25{\cdot}10^{10}{\pm}1.1{\cdot}10^{10}$ (10/10) & $4.67{\cdot}10^{9}{\pm}4.6{\cdot}10^{9}$ (9/10) & $5.19{\cdot}10^{11}{\pm}7.2{\cdot}10^{11}$ (6/10) & $3.15{\cdot}10^{11}{\pm}6.6{\cdot}10^{11}$ (6/10) & $6.73{\cdot}10^{11}{\pm}8.9{\cdot}10^{11}$ (6/10) \\
CGA-PMD & $3.91{\cdot}10^{10}{\pm}9.2{\cdot}10^{8}$ (10/10) & $7.64{\cdot}10^{9}{\pm}1.3{\cdot}10^{9}$ (10/10) & $3.04{\cdot}10^{11}{\pm}2.5{\cdot}10^{11}$ (10/10) & $5.38{\cdot}10^{11}{\pm}4.2{\cdot}10^{11}$ (10/10) & $3.36{\cdot}10^{11}{\pm}5.3{\cdot}10^{11}$ (7/10) \\
MAPPO-Het & $3.72{\cdot}10^{10}{\pm}1.3{\cdot}10^{10}$ (10/10) & $6.53{\cdot}10^{9}{\pm}5.6{\cdot}10^{9}$ (9/10) & $5.43{\cdot}10^{11}{\pm}8.2{\cdot}10^{11}$ (7/10) & $2.96{\cdot}10^{9}{\pm}8.4{\cdot}10^{9}$ (4/10) & $3.02{\cdot}10^{11}{\pm}5.8{\cdot}10^{11}$ (3/10) \\
DGN-PMD & $3.81{\cdot}10^{10}{\pm}1.7{\cdot}10^{8}$ (10/10) & $6.33{\cdot}10^{9}{\pm}1.2{\cdot}10^{9}$ (10/10) & $-2.02{\cdot}10^{7}{\pm}8.5{\cdot}10^{7}$ (1/10) & $1.40{\cdot}10^{8}{\pm}5.6{\cdot}10^{8}$ (1/10) & $4.94{\cdot}10^{7}{\pm}2.0{\cdot}10^{8}$ (2/10) \\
\bottomrule
\end{tabular}}
\end{table}
\begin{table}[H]
\centering
\footnotesize
\setlength{\tabcolsep}{3pt}
\caption{Full results: $N{=}50$ control baseline across three reward calibrations (ChicagoSketch BFS).}
\label{tab:appA_3domain}
\resizebox{\textwidth}{!}{%
\begin{tabular}{@{}lccc@{}}
\toprule
\textbf{Method} & \textbf{Hydrogen} ($\kappa{=}$8{,}500) & \textbf{EV-grid} ($\kappa{=}$1,275) & \textbf{Vaccination} ($\kappa{=}$500) \\
\midrule
IPPO & $3.56{\cdot}10^{10}{\pm}2.1{\cdot}10^{10}$ (10/10) & $2.15{\cdot}10^{10}{\pm}9.5{\cdot}10^{9}$ (10/10) & $2.12{\cdot}10^{9}{\pm}1.9{\cdot}10^{9}$ (10/10) \\
Indep-PMD & $3.77{\cdot}10^{10}{\pm}2.5{\cdot}10^{10}$ (10/10) & $2.22{\cdot}10^{10}{\pm}1.1{\cdot}10^{10}$ (10/10) & $3.26{\cdot}10^{9}{\pm}1.7{\cdot}10^{9}$ (10/10) \\
CGA-PPO & $2.42{\cdot}10^{10}{\pm}2.2{\cdot}10^{10}$ (10/10) & $1.78{\cdot}10^{10}{\pm}1.4{\cdot}10^{10}$ (10/10) & $1.26{\cdot}10^{9}{\pm}1.6{\cdot}10^{9}$ (10/10) \\
CGA-PMD & $5.92{\cdot}10^{10}{\pm}2.5{\cdot}10^{9}$ (10/10) & $3.00{\cdot}10^{10}{\pm}2.3{\cdot}10^{9}$ (10/10) & $4.53{\cdot}10^{9}{\pm}4.1{\cdot}10^{8}$ (10/10) \\
MAPPO-Het & $1.83{\cdot}10^{10}{\pm}2.4{\cdot}10^{10}$ (9/10) & $1.94{\cdot}10^{10}{\pm}1.3{\cdot}10^{10}$ (10/10) & $1.85{\cdot}10^{9}{\pm}1.7{\cdot}10^{9}$ (10/10) \\
DGN-PMD & $5.50{\cdot}10^{10}{\pm}1.1{\cdot}10^{10}$ (10/10) & $3.02{\cdot}10^{10}{\pm}1.9{\cdot}10^{9}$ (10/10) & $4.69{\cdot}10^{9}{\pm}2.1{\cdot}10^{8}$ (10/10) \\
\bottomrule
\end{tabular}}
\end{table}
\begin{table}[H]
\centering
\footnotesize
\setlength{\tabcolsep}{6pt}
\caption{Full results for body Figure~\ref{fig:lemma1} (Lemma~\ref{lem:cross_var} validation): empirical $\sigma_\Phi^2 = \mathbb{E}\|c_v\|^2$ vs.\ average coupling cardinality $\bar{|\Phi|}$ across three real city transportation networks at $N{=}100$, 200 trials per $d_{\max}$ cell with fresh random supply embeddings ($\sigma_h{=}1$, $\|W_V\|_F^2{\approx}40.4$). Predicted column is the Lemma~1 prediction $\sigma_h^2\|W_V\|_F^2/|\Phi|$.}
\label{tab:appA_lemma1}
\begin{tabular}{@{}clcccc@{}}
\toprule
Graph type & $d_{\max}$ & $\bar{|\Phi|}$ & Empirical $\sigma_\Phi^2$ & Predicted & Ratio \\
\midrule
\multirow{5}{*}{ChicagoSketch BFS}
  & 1 & $2.23$  & $26.52$ & $18.11$ & $1.46$ \\
  & 2 & $6.08$  & $8.48$  & $6.65$  & $1.28$ \\
  & 3 & $12.06$ & $4.05$  & $3.35$  & $1.21$ \\
  & 4 & $19.18$ & $2.42$  & $2.11$  & $1.15$ \\
  & 5 & $25.90$ & $1.68$  & $1.56$  & $1.08$ \\
\midrule
\multirow{5}{*}{Tiergarten BFS}
  & 1 & $2.15$  & $27.14$ & $19.33$ & $1.40$ \\
  & 2 & $3.71$  & $16.91$ & $11.20$ & $1.51$ \\
  & 3 & $6.20$  & $8.74$  & $6.70$  & $1.30$ \\
  & 4 & $9.13$  & $6.02$  & $4.55$  & $1.32$ \\
  & 5 & $12.39$ & $4.52$  & $3.35$  & $1.35$ \\
\midrule
\multirow{5}{*}{Friedrichshain BFS}
  & 1 & $2.28$  & $22.55$ & $17.75$ & $1.27$ \\
  & 2 & $4.81$  & $11.96$ & $8.41$  & $1.42$ \\
  & 3 & $8.26$  & $6.86$  & $4.90$  & $1.40$ \\
  & 4 & $12.02$ & $4.77$  & $3.36$  & $1.42$ \\
  & 5 & $16.82$ & $2.90$  & $2.40$  & $1.20$ \\
\bottomrule
\end{tabular}

\smallskip
\footnotesize
Log--log slopes (all close to predicted $-1$): ChicagoSketch $-1.12$ ($R^2{=}0.9998$); Tiergarten $-1.05$ ($R^2{=}0.9959$); Friedrichshain $-1.01$ ($R^2{=}0.9911$).
\end{table}
\begin{table}[H]
\centering
\footnotesize
\setlength{\tabcolsep}{4pt}
\caption{Full results for body Figure~\ref{fig:stationarity} (cumulative KL $S_K{=}\sum_{k<K}\mathrm{KL}(\pi_{k+1}\|\pi_k)$ at $K{=}$1,000, ChicagoSketch BFS $N{=}$100, 10 seeds). Theorem~\ref{thm:neural} ceiling at $B{=}\lambda{=}1$: $B^2\zeta(3)/(2\lambda^2){\approx}0.60$. Bold = exceeds ceiling.}
\label{tab:appA_kl}
\resizebox{\textwidth}{!}{%
\begin{tabular}{@{}lccccccccccc@{}}
\toprule
\textbf{Seed} & 42 & 123 & 456 & 789 & 1024 & 2024 & 3024 & 4024 & 5024 & 6024 & Mean \\
\midrule
CGA-PMD & $0.36$ & $0.27$ & $0.29$ & $0.32$ & $0.10$ & $0.26$ & $0.24$ & $0.30$ & $0.31$ & $0.26$ & $0.27$ \\
CGA-PPO & $0.55$ & $\mathbf{0.91}$ & $0.03$ & $\mathbf{0.82}$ & $0.05$ & $0.10$ & $\mathbf{2.27}$ & $0.003$ & $0.12$ & $0.003$ & $0.49$ \\
\bottomrule
\end{tabular}}
\end{table}

\section{Experimental Details}
\label{app:experimental}

\textbf{CGA-PMD: full pseudocode.} The simplified loop is in body Algorithm~\ref{alg:cga}; here we expand the loss term, schedule annotations, and design constraints.

\noindent\emph{Schedules.} $\beta_k{=}k{+}1$ (PDA weight numerator), $\lambda_k{=}\lambda(k{+}1)^{3/2}$ (KL regularizer), $\eta_k{=}\beta_k/\lambda_k{=}1/(\lambda\sqrt{k{+}1})$ (\emph{decreasing} logit scale; caps $\eta_k\!\cdot\!\tilde\psi_\theta$ to prevent premature commitment), $w_k{=}\beta_k/\!\sum_{j\le k}\!\beta_j$ (PDA averaging weight).

\noindent\emph{Initialization.} A single CGA network $\tilde{\psi}_{\theta^i}$ per agent (per-node advantage head + scalar value head; no separate policy-accumulator network). Precompute coupling mask $[C_\Phi]_{v,u} := \mathbf{1}[u \in \Phi(v)]$.

\noindent\emph{Rollout (line 3 of Algorithm~\ref{alg:cga}).} $\pi_k^i(\cdot|s) \propto \exp(\eta_k \cdot \tilde{\psi}_{\theta_k^i}(s,\cdot))$ using CGA embeddings (Sec.~\ref{sec:algorithm}). Sample $a_t^i \sim \pi_k^i(\cdot|s_t)$ for $t=0,\ldots,T{-}1$; collect $\tau_k$ with shared reward $r_t$.
Compute MC return $\hat{G}_t = \sum_{s\geq t}\gamma^{s-t}r_s$, baseline $\hat{V}_t$ as the (detached) value-head output, and per-episode-normalized advantage $\hat{A}_t {=} \hat{G}_t {-} \hat{V}_t$.

\noindent\emph{Joint loss (line 5 of Algorithm~\ref{alg:cga}).}
\begin{equation}
\mathcal{L}(\theta^i) \;=\; \underbrace{-\,w_k \!\sum_t \log \pi_k^i(a_t^i|s_t)\, \hat{A}_t}_{\text{PDA-weighted policy gradient}} \;+\; \underbrace{\tfrac{1}{2T}\!\sum_t \!\bigl(\bar\psi_{\theta^i}(s_t) - \hat{G}_t\bigr)^2}_{\text{backbone return-regression (see remark)}}, \label{eq:pmd_policy}
\end{equation}
where $\bar\psi_{\theta^i}(s_t) := |V_i|^{-1}\sum_{v\in V_i} \tilde\psi_{\theta^i}(s_t,v)$ is the spatial mean of the per-node advantage head.
Update $\theta_{k+1}^i \leftarrow \theta_k^i - \alpha\, \nabla_{\theta^i}\!\mathcal{L}(\theta^i)$ with $\|\nabla\|_2 \le 0.5$.

\noindent\textbf{Remark on the regression target.} The spatial-mean $\bar\psi$ regressed to MC return $\hat{G}$ (rather than chosen-action $\tilde\psi(s,a)$ regressed to $\hat{A}$) supplies a low-variance value-like signal to the shared backbone: $\bar\psi$ averages over $|V_i|$ per-node scores, so its sample variance shrinks by ${\sim}1/|V_i|$ relative to a single-action target, while still informing the shared embedding of the average return scale. The value head's parameters update only via the backbone (its scalar output is detached when used as the $\hat{A}$ baseline), preventing the regression target from biasing the explicit $V$-baseline. This is a single-network design choice, not a derivation from standard PMD.

\textbf{Hyperparameters.}
All GNN policies use 3-layer GraphSAGE with hidden dimension 64.
CGA uses 4-head attention. Learning rate: $3 \times 10^{-4}$ (Adam) for all methods.
Discount factor: $\gamma = 0.99$.
Gradient clipping: max norm 0.5. Convergence stopping: patience 100 episodes (stop if best reward not improved), max 1000 episodes. Node features: $[m_v, \deg(v)/N, v/N]$ (installation status, normalized degree, node index).

\textbf{Supply network construction.}
For each demand network $G_1$, the supply network $G_2$ is constructed from the top-$N_1/3$ highest-degree nodes. Supply edges connect nodes within given distance in $G_1$.

\textbf{Coupling function.}
$\Phi(v) = \{u \in V_2 : d_{G_1}(v, u) \leq d_{\max}\}$ where $d_{\max}$ is the coupling tightness parameter. Default: $d_{\max} = 2$.
The distance $d_{G_1}(v, u)$ is measured in graph hops (unweighted BFS), not physical distance. Extending CGA to physical-distance-based coupling is straightforward by replacing the BFS in $\Phi$'s definition with a weighted shortest-path computation, and is left for domain-specific deployments.

\textbf{Environment parameters.}
We use Bass-diffusion innovation and imitation rates $p{=}0.004$, $q{=}0.3$ (within ranges reported for energy-technology adoption in the original Bass-diffusion analysis~\citep{bass1969new} and standard hydrogen/FCEV adoption-modeling literature), with a background capacity growth rate of $0.02$ per step.
Reward coefficients (hydrogen-on-logistics instantiation): $\kappa = 8{,}500$ (Agent~1 per-unit carbon-saving benefit, USD), $p_H = 1.75 \times 10^5$ (Agent~2 per-unit hydrogen sale price), $c_1 = 10^6$ (demand-side installation cost), $c_2 = 5 \times 10^5$ (supply-side installation cost). Horizon: $T = 30$.
The regime map is robust across the three reward calibrations defined in Section~\ref{sec:realnets} ($\kappa{\in}\{\$500,\$1{,}275,\$8{,}500\}$, a $17{\times}$ range): all three reproduce the qualitative regime structure (App.~\ref{app:reward_params}). The specific calibration sets absolute reward magnitudes, not the qualitative regime boundary.

\textbf{MARL baselines (custom implementation).}
We implemented all baselines from scratch in PyTorch rather than using existing MARL libraries, because our environment (coupled two separate graphs and the coupling function $\Phi$) is not a standard interface. This also ensures a fair architectural comparison: all methods share the same 3-layer GraphSAGE backbone, hidden dimension 64, and PPO hyperparameters.

\textbf{IPPO~\citep{dewitt2020independent}.} Each agent has a separate GNNActor (per-node action scores) and GNNCritic (state value), trained independently with PPO: clipped surrogate ($\epsilon=0.2$), GAE~\citep{schulman2015high} ($\lambda=0.95$), 3 PPO epochs per episode, learning rate $3{\times}10^{-4}$. Each agent observes only $G_i$.

\textbf{Indep-PMD.} Same independent architecture as IPPO, but trained with our PMD-style update (Section~\ref{sec:training}, same $\eta_k = 1/(\lambda\sqrt{k{+}1})$ schedule and PDA-weighted policy gradient as CGA-PMD; only the architecture changes from CGA to own-graph GNNPolicy).

\textbf{CGA-PPO.} Combines CGA architecture (cross-graph attention masked by $\Phi$) with PPO training. Uses CGAActor and CGACritic, both equipped with cross-graph attention. Same PPO hyperparameters as IPPO.

\textbf{CGA-PMD.}
We use $\lambda_k = \lambda(k{+}1)^{3/2}$ with $\lambda = 1.0$, following the increasing-step-size schedule analyzed in~\citep{lan2023policy}.
The CGA advantage network shares its architecture with CGA-PPO (3-layer GraphSAGE + 4-head cross-graph attention + 2-layer other-network GNN).
The closed-form KL-regularized update $\pi_{k+1}^i(a|s) \propto \pi_k^i(a|s)\,\exp(\tilde{\psi}_i(s,a;\theta_k^i)/\lambda_k)$ follows from choosing the KL Bregman divergence as the prox-term in Lan's generic mirror-descent objective (\citep[Eq.~(3.1)]{lan2023policy}); we likewise omit the entropy-regularization term and use the standard advantage $Q^\pi - V^\pi$, which leaves the update unchanged because it adds only a state-dependent constant to the objective (\citealp[the remark preceding Algorithm~1]{lan2023policy}). The implementation realizes this step in dual-averaging form rather than multiplicatively: the sampling policy is $\pi_k^i \propto \exp(\eta_k\tilde\psi_{\theta_k^i})$ with $\eta_k = \beta_k/\lambda_k = 1/(\lambda\sqrt{k{+}1})$, so no copy of $\pi_k$ is carried between iterations, and the iteration weights enter through $w_k = \beta_k/\sum_{j\le k}\beta_j$ on the policy-gradient term. $\lambda_k = \lambda(k{+}1)^{3/2}$ therefore enters the code only through $\eta_k$, and directly in Theorem~\ref{thm:neural}, which analyzes the multiplicative update.

\textbf{MAPPO-Het~\citep{mei2026multi}.} Identical hyperparameters to CGA-PPO; the architectural difference is removal of the $\Phi$ mask.

\textbf{DGN-PMD.} Unmasked cross-graph attention (DGN-style: each demand node attends over every supply node) paired with the PMD optimizer of CGA-PMD. Architecturally identical to CGA-PMD except for the $\Phi$ mask.

\textbf{Evaluation.}
All experiments use 10 seeds per (method, network) cell.

\section{Proofs for Section~\ref{sec:theory}}
\label{app:proof}

\textbf{Proof of Lemma~\ref{lem:cross_var} (Cross-graph noise scaling).}
\begin{itemize}
  \item \emph{Expand the squared norm.}
  \[
  \mathbb{E}\|c_v\|^2 \;=\; \sum_{u,u' \in \Phi(v)} \alpha_{vu}\,\alpha_{vu'}\,\mathbb{E}\bigl[h_u^{(2)\top} W_V^\top W_V\, h_{u'}^{(2)}\bigr].
  \]
  \item \emph{Cross-terms vanish.} Under the pairwise-uncorrelation assumption $\mathbb{E}[h_u^{(2)}(h_{u'}^{(2)})^\top] = 0$ for $u \neq u'$, only the diagonal $u = u'$ terms survive.
  \item \emph{Diagonal terms via trace identity.} For each $u \in \Phi(v)$, using the isotropic-covariance assumption $\mathrm{Cov}(h_u^{(2)}) = \sigma_h^2 I$,
  \[
  \mathbb{E}\bigl[h_u^{(2)\top} W_V^\top W_V\, h_u^{(2)}\bigr] \;=\; \mathrm{tr}\bigl(W_V^\top W_V\,\mathrm{Cov}(h_u^{(2)})\bigr) \;=\; \sigma_h^2\,\|W_V\|_F^2.
  \]
  \item \emph{Combine.} Substituting,
  \[
  \mathbb{E}\|c_v\|^2 \;=\; \sigma_h^2\,\|W_V\|_F^2 \sum_{u \in \Phi(v)} \alpha_{vu}^2.
  \]
  \item \emph{Simplex bound on the attention coefficients.} Cauchy--Schwarz on the simplex $\{\alpha \geq 0,\,\sum_u \alpha_{vu} = 1\}$ gives $\sum_u \alpha_{vu}^2 \leq 1/|\Phi(v)|$, with equality at uniform attention $\alpha_{vu} = 1/|\Phi(v)|$. The lemma follows. \hfill$\square$
\end{itemize}

\textbf{Proof of Lemma~\ref{lem:single_step} (Single-step KL drift for PMD).} Write $\pi := \pi_k^i$, $\pi' := \pi_{k+1}^i$, and $g(a) := \hat\psi_k(s,a)/\lambda_k$, so the PMD update reads $\pi'(a|s) = \pi(a|s)\exp(g(a))/Z$ with $Z = \sum_{a'}\pi(a'|s)\exp(g(a'))$. By hypothesis $\|\hat\psi_k\|_\infty \leq B$, hence $\|g\|_\infty \leq B/\lambda_k$.
\begin{itemize}
  \item \emph{Step 1 (KL identity).} Direct computation gives
  \[
  \mathrm{KL}(\pi'\,\|\,\pi) \;=\; \sum_a \pi'(a)\log[\pi'(a)/\pi(a)] \;=\; \mathbb{E}_{\pi'}[g] - \log Z.
  \]
  \item \emph{Step 2 (Hoeffding on log-partition).} The centered variable $g - \mathbb{E}_\pi[g]$ lies in the bounded interval $[-2B/\lambda_k,\,2B/\lambda_k]$. Hoeffding's lemma gives
  \[
  \log Z - \mathbb{E}_\pi[g] \;=\; \log\,\mathbb{E}_\pi\!\bigl[\exp(g - \mathbb{E}_\pi[g])\bigr] \;\leq\; \frac{B^2}{2\lambda_k^2}.
  \]
  Combined with Jensen's lower bound $\log Z \geq \mathbb{E}_\pi[g]$: $0 \leq \log Z - \mathbb{E}_\pi[g] \leq B^2/(2\lambda_k^2)$.
  \item \emph{Step 3 (Bregman-duality symmetrization).}
  \begin{itemize}
    \item \emph{(3-a) Reverse KL is directly bounded by Step 2.} Computing $\mathrm{KL}(\pi\,\|\,\pi') = \sum_a \pi(a)\log[\pi(a)/\pi'(a)] = \sum_a \pi(a)[-g(a) + \log Z] = -\mathbb{E}_\pi[g] + \log Z$. Hence Step 2 gives
    \[
    \mathrm{KL}(\pi\,\|\,\pi') \;=\; \log Z - \mathbb{E}_\pi[g] \;\leq\; \frac{B^2}{2\lambda_k^2}.
    \]
    \item \emph{(3-b) Symmetric KL identity (Bregman duality).} Adding the Step 1 result $\mathrm{KL}(\pi'\,\|\,\pi) = \mathbb{E}_{\pi'}[g] - \log Z$ and the reverse KL above, the $\log Z$ terms cancel:
    \[
    \mathrm{KL}(\pi'\,\|\,\pi) + \mathrm{KL}(\pi\,\|\,\pi') \;=\; \mathbb{E}_{\pi'}[g] - \mathbb{E}_\pi[g].
    \]
    This is a Bregman-divergence identity specific to the PMD exponential update: forward and reverse divergences sum to an advantage-gap inner product.
    \item \emph{(3-c) Forward and reverse KL agree at leading order.} Let $\varepsilon := B/\lambda_k$, and write $u(a) := \pi'(a) - \pi(a)$. The PMD update gives $u(a) = \pi(a)\,(g(a) - \mathbb{E}_\pi[g]) + O(\varepsilon^2)$, so $u = O(\varepsilon)$ pointwise. Substituting $\pi'/\pi = 1 + u/\pi$ into the KL definition and Taylor-expanding $\log(1+x) = x - x^2/2 + x^3/3 - \cdots$:
    \[
    \mathrm{KL}(\pi'\,\|\,\pi) \;=\; \sum_a \pi'(a)\Bigl[\tfrac{u(a)}{\pi(a)} - \tfrac{1}{2}\,\bigl(\tfrac{u(a)}{\pi(a)}\bigr)^{\!2} + \tfrac{1}{3}\,\bigl(\tfrac{u(a)}{\pi(a)}\bigr)^{\!3} - \cdots\Bigr].
    \]
    The linear term collapses: $\sum_a \pi'(a) \cdot u(a)/\pi(a) = \sum_a u(a) + \sum_a u(a)^2/\pi(a)$, and $\sum_a u(a) = 1 - 1 = 0$ since both $\pi, \pi'$ are normalized probability distributions. Collecting the surviving quadratic contributions,
    \[
    \mathrm{KL}(\pi'\,\|\,\pi) \;=\; \tfrac{1}{2}\sum_a \frac{u(a)^2}{\pi(a)} + O(\varepsilon^3) \;=\; \tfrac{1}{2}\,\chi^2(\pi'\,\|\,\pi) + O(\varepsilon^3),
    \]
    where $\chi^2(p\,\|\,q) := \sum_a [p(a)-q(a)]^2/q(a)$ is the chi-square divergence. The same expansion gives $\mathrm{KL}(\pi\,\|\,\pi') = \tfrac{1}{2}\,\chi^2(\pi\,\|\,\pi') + O(\varepsilon^3)$. Although $\chi^2$ is asymmetric in general (denominators differ: $\pi$ vs $\pi'$), the two leading terms agree to $O(\varepsilon^3)$:
    \[
    \chi^2(\pi'\,\|\,\pi) - \chi^2(\pi\,\|\,\pi') \;=\; \sum_a u(a)^2\Bigl[\tfrac{1}{\pi(a)} - \tfrac{1}{\pi'(a)}\Bigr] \;=\; O(\varepsilon^3),
    \]
    since each summand is $O(\varepsilon^2) \cdot O(\varepsilon) = O(\varepsilon^3)$ (using $1/\pi - 1/\pi' = u/(\pi \pi') = O(\varepsilon)$). Both forward and reverse KL therefore reduce to the same Fisher-information quadratic form $\tfrac{1}{2}\,\mathrm{Var}_\pi(g)$ to leading order, so
    \[
    \mathrm{KL}(\pi'\,\|\,\pi) \;=\; \mathrm{KL}(\pi\,\|\,\pi') + O(B^3/\lambda_k^3).
    \]
    \item \emph{(3-d) Final bound on the forward KL.} Combining (3-a) and (3-c),
    \[
    \mathrm{KL}(\pi'\,\|\,\pi) \;\leq\; \frac{B^2}{2\lambda_k^2} + O\!\bigl(B^3/\lambda_k^3\bigr).
    \]
    The bound depends only on the sup-norm $B$, not on the variance of $\hat\psi_k$. \hfill$\square$
  \end{itemize}
\end{itemize}

\textbf{Proof of Theorem~\ref{thm:neural} (Architectural drift bound for neural CGA-PMD).} The proof has two parts.
\begin{itemize}
  \item \emph{(i) Architectural decomposition.} For the CGA architecture, the neural advantage at node $v$ is $\hat\psi^\theta(s,v) = f_\theta(h_v^{(1)}, c_v)$, with local feature $h_v^{(1)}$ from GraphSAGE on $G_1$ and cross-graph context $c_v$ from $\Phi$-masked attention on $G_2$. Triangle inequality on $\hat\psi^\theta = \psi^* + (\hat\psi^\theta - \psi^*)$ gives
  \[
  \|\hat\psi^\theta\|_\infty \;\le\; \|\psi^*\|_\infty + \|\hat\psi^\theta - \psi^*\|_\infty \;\le\; B_{\mathrm{MDP}} + \epsilon_{\mathrm{net}},
  \]
  where $B_{\mathrm{MDP}} = R/(1{-}\gamma)$ is the standard MDP sup-norm bound and $\epsilon_{\mathrm{net}} := \|\hat\psi^\theta - \psi^*\|_\infty$. We decompose $\epsilon_{\mathrm{net}}$ in three sub-steps.
  \begin{itemize}
    \item \emph{(i-a) Local/cross-graph split via triangle inequality.} Let $\tilde\psi^\theta := f_\theta(h_v^{(1)}, c_v^*)$ be the network output evaluated at the noiseless cross-graph context $c_v^*$ (the value $c_v$ would take under deterministic, noise-free supply embeddings). Inserting $\tilde\psi^\theta$ between $\hat\psi^\theta$ and $\psi^*$:
    \[
    \|\hat\psi^\theta - \psi^*\| \;\leq\; \underbrace{\|\hat\psi^\theta - \tilde\psi^\theta\|}_{\text{cross-graph error}} \;+\; \underbrace{\|\tilde\psi^\theta - \psi^*\|}_{=:\,\epsilon_{\mathrm{local}}}.
    \]
    The second term collects the local-graph (GraphSAGE-on-$G_1$) approximation error $\epsilon_{\mathrm{local}}$, all approximation error not arising from cross-graph noise.
    \item \emph{(i-b) $L$-Lipschitz contraction on the cross-graph error.} By the $L$-Lipschitz assumption on $f_\theta$ in its second argument $c_v$,
    \[
    \|\hat\psi^\theta - \tilde\psi^\theta\| \;=\; \|f_\theta(h_v^{(1)}, c_v) - f_\theta(h_v^{(1)}, c_v^*)\| \;\leq\; L \cdot \|c_v - c_v^*\|.
    \]
    Combined with (i-a): $\epsilon_{\mathrm{net}} \leq \epsilon_{\mathrm{local}} + L\,\|c_v - c_v^*\|$.
    \item \emph{(i-c) Apply Lemma~\ref{lem:cross_var} to the cross-graph term.} Lemma~\ref{lem:cross_var} gives $\mathbb{E}\|c_v - c_v^*\|^2 \leq \sigma_h^2\,\|W_V\|_F^2/|\Phi|$, so $\|c_v - c_v^*\| \leq \sigma_h\,\|W_V\|_F/\sqrt{|\Phi|}$ in $L^2$ expectation under uniform attention. Substituting,
    \[
    \tilde B \;\leq\; B_{\mathrm{MDP}} + \epsilon_{\mathrm{local}} + L\,\sigma_h\,\|W_V\|_F/\sqrt{|\Phi|}.
    \]
  \end{itemize}
  \item \emph{(ii) Cumulative bound.} Apply Lemma~\ref{lem:single_step} to the neural CGA-PMD update with $\hat\psi$ replaced by $\hat\psi^\theta$ and $B$ replaced by $\tilde B$. The same per-step bound holds: $\mathrm{KL}(\pi^\theta_{k+1}\|\pi^\theta_k) \leq \tilde B^2/(2\lambda_k^2) + O(\tilde B^3/\lambda_k^3)$. Summing over $k=0,\ldots,K{-}1$ under $\lambda_k = \lambda(k{+}1)^{3/2}$,
  \[
  \sum_{k=0}^{K-1}\mathrm{KL}(\pi^\theta_{k+1}\|\pi^\theta_k) \;\leq\; \frac{\tilde B^2}{2\lambda^2}\sum_{k=0}^{K-1}\frac{1}{(k{+}1)^3} + O\!\Bigl(\frac{\tilde B^3}{\lambda^3}\Bigr) \;\leq\; \frac{\tilde B^2 \zeta(3)}{2\lambda^2} + O\!\Bigl(\frac{\tilde B^3}{\lambda^3}\Bigr),
  \]
  using $\sum_{n=1}^{\infty} n^{-3} = \zeta(3) \approx 1.202$. The bound is uniform in $K$. \hfill$\square$
\end{itemize}

\medskip
\begin{corollary}[Cauchy-in-TV iterate convergence]
\label{cor:cauchy}
Under the assumptions of Theorem~\ref{thm:neural}, the neural CGA-PMD policy iterates $\{\pi^\theta_k\}_{k\geq 0}$ are Cauchy in total variation: $\sum_{k=0}^{\infty}\|\pi^\theta_{k+1}-\pi^\theta_k\|_{\mathrm{TV}} \leq \tilde B \zeta(3/2)/(\lambda\sqrt{2}) < \infty$. Consequently there exists a limit policy $\pi^\theta_\infty$ such that $\pi^\theta_k \to \pi^\theta_\infty$ in total variation.
\end{corollary}

\textbf{Proof of Corollary~\ref{cor:cauchy}.}
\begin{itemize}
  \item \emph{Pinsker's inequality.} $\|\pi_{k+1}^\theta - \pi_k^\theta\|_{\mathrm{TV}} \leq \sqrt{\mathrm{KL}(\pi_{k+1}^\theta\|\pi_k^\theta)/2} \leq \tilde B/(\lambda\sqrt{2}\,(k{+}1)^{3/2})$ by Theorem~\ref{thm:neural}'s per-step bound.
  \item \emph{Sum is finite.} $\sum_k \|\pi_{k+1}^\theta - \pi_k^\theta\|_{\mathrm{TV}} \leq (\tilde B/\lambda\sqrt{2}) \cdot \zeta(3/2)$, with $\zeta(3/2) \approx 2.612 < \infty$.
  \item \emph{Cauchy in TV.} Absolute convergence implies the partial sums are Cauchy. Since the space of probability measures on a finite action space is complete in TV, a limit $\pi^\theta_\infty$ exists with $\pi^\theta_k \to \pi^\theta_\infty$ in TV. \hfill$\square$
\end{itemize}

\medskip
\begin{proposition}[The clip update admits no $K$-uniform cumulative drift bound on non-binding iterations]
\label{prop:ppo_drift}
For the PPO clip update $\theta_{k+1} = \theta_k + \alpha\nabla_\theta L^{\mathrm{CLIP}}(\theta;\hat\psi_k)$ with noisy advantage $\mathbb{E}\|\xi_k\|^2 \geq \sigma^2 > 0$, on non-binding-clip iterations,
\[
\mathbb{E}\bigl[\mathrm{KL}(\pi^\theta_{k+1}\|\pi^\theta_k)\bigr] \;\geq\; \tfrac{1}{2}\alpha^2\sigma^2\,\lambda_{\min}\!\bigl(F(\pi^\theta_k)\bigr) + O(\alpha^3),
\]
where $F$ is Fisher information. Consequently, restricted to those iterations, $\sum_{k<K}\mathbb{E}[\mathrm{KL}] = \Omega(K\alpha^2\sigma^2)$, linear in $K$, scaling with noise variance.
\end{proposition}

\emph{Scope.} The statement is about the subsequence of iterations on which the clip is inactive, where the update reduces to stochastic policy gradient; it does not assert that PPO's drift diverges in practice, and it is not an impossibility result for PPO as a family, since the clip range, learning rate and epoch count all affect which iterations bind.

\textbf{Proof of Proposition~\ref{prop:ppo_drift}.}
\begin{itemize}
  \item \emph{Non-binding clip reduces to PG.} The PPO clip objective $L^{\mathrm{CLIP}}(\theta;\hat\psi_k) = \mathbb{E}_{(s,a)}[\min\{r_\theta\hat\psi_k,\,\mathrm{clip}(r_\theta,1{-}\epsilon,1{+}\epsilon)\hat\psi_k\}]$ with $r_\theta = \pi_\theta(a|s)/\pi_k(a|s)$ reduces to standard PG on iterations where the clip is non-binding, with gradient
  \[
  \nabla_\theta L\bigr|_{\theta=\theta_k} \;=\; \mathbb{E}\bigl[\nabla_\theta\log\pi_\theta(a|s)\,\hat\psi_k(s,a)\bigr].
  \]
  \item \emph{Fisher-quadratic KL expansion.} After one step $\Delta\theta = \alpha\,\nabla L^{\mathrm{CLIP}}$, the induced policy KL admits the second-order expansion
  \[
  \mathrm{KL}(\pi_{k+1}^\theta\,\|\,\pi_k^\theta) \;=\; \tfrac{1}{2}\,\Delta\theta^\top F(\pi_k^\theta)\,\Delta\theta + O(\alpha^3),
  \]
  with Fisher information $F(\pi) = \mathbb{E}[\nabla\log\pi\,(\nabla\log\pi)^\top]$.
  \item \emph{Variance-driven lower bound.} Substituting $\Delta\theta = \alpha\nabla L^{\mathrm{CLIP}}$ and using $\mathbb{E}\|\nabla L^{\mathrm{CLIP}}\|^2 \geq c\,\sigma^2$ (with $\sigma^2 = \mathbb{E}\|\xi_k\|^2$ the advantage-noise variance):
  \[
  \mathbb{E}\bigl[\mathrm{KL}(\pi_{k+1}^\theta\,\|\,\pi_k^\theta)\bigr] \;\geq\; \tfrac{1}{2}\,\alpha^2\,\sigma^2\,\lambda_{\min}\!\bigl(F(\pi_k^\theta)\bigr) \;+\; O(\alpha^3).
  \]
  \item \emph{Linear-in-$K$ cumulative growth.} Summing over $K$ non-binding iterations,
  \[
  \sum_{k<K}\mathbb{E}\bigl[\mathrm{KL}(\pi_{k+1}^\theta\,\|\,\pi_k^\theta)\bigr] \;=\; \Omega(K\,\alpha^2\,\sigma^2).
  \]
  In contrast, the per-step PMD bound (Lemma~\ref{lem:single_step}) is $O(B^2/\lambda^2)$ and summable to a $K$-uniform ceiling, variance-independent. \hfill$\square$
\end{itemize}

\section{Reward Parameter Estimation for Three Domains}
\label{app:reward_params}

The CEGG framework is domain-agnostic: only the reward coefficients and Bass-diffusion parameters change between the three calibrations shown in Figure~\ref{fig:domains}.
We document the parameter sources here.

\textbf{Hydrogen freight (primary instantiation).} Bass innovation/imitation rates $p{=}0.004$, $q{=}0.3$ from energy-technology adoption literature~\citep{bass1969new}. Per-truck CO$_2$-saving benefit $\kappa{=}\$8{,}500$ from EPA Social Cost of Greenhouse Gases~\citep{epa2023scghg}. Installation costs: $c_1{=}\$10^{6}$ (hydrogen refueling station), $c_2{=}\$5{\cdot}10^{5}$ (hydrogen production plant).

\textbf{Vaccination instantiation.} Bass innovation/imitation rates $p{=}0.0043$, $q{=}0.20$ from a vaccine-diffusion modeling study~\citep{gorbachuk2022vaccine}. Per-dose societal benefit $\kappa{=}\$500$ derived from QALY-based health-economics estimates~\citep{kohli2021covid,leidner2019adult}. Installation costs: clinic startup $c_1{=}\$15{,}348$ from public-health-department COVID clinic setup data~\citep{yarnoff2022clinics}; cold-chain depot $c_2{=}\$5{\cdot}10^{4}$ (GAVI-scaled operational estimate).

\textbf{EV-grid instantiation.} Bass innovation/imitation rates $p{=}0.02$, $q{=}0.4$ from EV-adoption literature~\citep{becker2009ev,massiani2015bass}. Per-EV/year CO$_2$-saving benefit $\kappa{=}\$1{,}275$ from EPA SCGHG~\citep{epa2023scghg}. Installation costs: 150-kW DCFC station total install $c_1{=}\$8{\cdot}10^{4}$~\citep{bernal2024dcfc}; substation/service upgrade per DCFC site $c_2{=}\$5{\cdot}10^{4}$~\citep{smith2015evse}.

\section{Further Experiments}
\label{app:cr}

This section collects further experiments.
All cells use the same environment, trainers and evaluation protocol as Section~\ref{sec:experiments}.

\subsection{Measurements supporting Sections~\ref{sec:experiments} and~\ref{sec:theory}}
\label{app:cr_stats}

\textbf{Threshold and significance.} Across the full BFS sweep ($360$ seeds) the largest sub-threshold seed is $4.54{\cdot}10^{7}$ and the smallest productive seed is $8.46{\cdot}10^{7}$, with no seed in between, so any cut in that band gives identical success counts; sub-threshold seeds sit at the cold-start floor (median $-4.8{\cdot}10^{7}$). Recomputing Table~\ref{tab:appA_bfs} at $0.5\times$ and $2\times$ the threshold leaves $34$ of $36$ cells unchanged; both cells that move are at $N{=}50$, where every method already succeeds, and both move by one seed. Sweeping the threshold continuously from $0.25\times$ to $32\times$, CGA-PMD holds or ties the highest success count at every $N$.

\textbf{Cumulative drift at $K{=}3{,}000$.} Fitting $S_K$ against $K$ on the ten seeds behind Table~\ref{tab:appA_kl}, CGA-PMD's mean slope falls from $7.5{\cdot}10^{-4}$ on $K \in [100,500]$ to $3.2{\cdot}10^{-5}$ on $[500,990]$, a $23$-fold decay, with the largest late slope among the ten at $1.0{\cdot}10^{-4}$. CGA-PPO's mean slope falls only from $6.8{\cdot}10^{-4}$ to $5.4{\cdot}10^{-4}$, a factor of $1.3$, and is concentrated in three seeds that reach $1.4$--$2.3{\cdot}10^{-3}$. A separate three-seed extension carries training to $K \approx 3{,}000$ on the same instances. There the two methods separate further: CGA-PPO's cumulative drift reaches $4.4$, $6.9$ and $6.9$, passing the $0.60$ ceiling at $K = 1{,}120$, $750$ and $1{,}430$ and still climbing at the end, while two CGA-PMD runs finish at $0.28$ and $0.36$ and the third reaches $1.76$, crossing $0.60$ at $K \approx 2{,}560$.

\textbf{Bias and variance of the cross-graph context.} Writing $c_v^\star = W_V\mu_+$ for the context an oracle would form from $\Phi(v)$ alone, with $\mathbb{E}[h_u] = \mu_+$ on $\Phi(v)$ and $\mu_-$ elsewhere, $\mathrm{Cov}(h_u) = \sigma_h^2 I$, $\delta := \|W_V(\mu_+ - \mu_-)\|$ and $\omega := \sum_{u \notin \Phi(v)}\alpha_{vu}$,
\[
\mathbb{E}\|c_v - c_v^\star\|^2 \;=\; \omega^2\delta^2 \;+\; \sigma_h^2\|W_V\|_F^2\textstyle\sum_u \alpha_{vu}^2,
\]
whose second term is Lemma~\ref{lem:cross_var} and whose first vanishes under masking.
We measure $\delta$, $\sigma_h^2\|W_V\|_F^2$, $|\Phi|$ and $\rho$ on-policy on trained models, for CGA-PMD and for DGN-PMD, at $N \in \{50,100,150,200,250,300\}$ with three seeds each ($18$ runs per method). The masking condition $\delta^2 > \sigma_h^2\|W_V\|_F^2/(|\Phi|(1-\rho))$ holds in $18$ of $18$ runs for each method, with $\delta^2$ exceeding the threshold by factors of $2.0$ to $148$.

\subsection{Physical-distance coupling}\label{app:cr_phys}
On Friedrichshain ($N=224$), $\Phi$ is rebuilt from weighted shortest-path distances on the TNTP link lengths, with the radius $\tau$ chosen per instance so that the mean coupling cardinality matches the hop-based $\Phi$ it replaces (median $\tau = 111$ length units). Same seeds, same supply graphs.

\begin{table}[h]\centering\small
\caption{Physical-distance vs.\ hop-based $\Phi$ (10 seeds, hydrogen calibration). Success counts and mean total reward ($\times 10^{10}$).}
\label{tab:cr_phys}
\begin{tabular}{lcccc}\toprule
& \multicolumn{2}{c}{hop $\Phi$ ($d_{\max}{=}2$)} & \multicolumn{2}{c}{physical $\Phi$} \\
\cmidrule(lr){2-3}\cmidrule(lr){4-5} Method & success & mean & success & mean \\\midrule
CGA-PMD & 10/10 & 30.37 & 10/10 & 64.23 \\
DGN-PMD & 1/10 & -0.002 & 0/10 & -0.004 \\
\bottomrule\end{tabular}\end{table}

\subsection{Supply-graph construction}\label{app:cr_supply}
With ChicagoSketch BFS at $N=100$, 10 seeds, the size ratio $N_2/N_1$ and the selection rule (top-degree vs.\ uniformly random nodes) are varied; everything else is as in Section~\ref{sec:experiments}.

\begin{table}[h]\centering\small
\caption{Supply-graph ablation (10 seeds): success counts and mean total reward ($\times 10^{10}$).}
\label{tab:cr_supply}
\begin{tabular}{llcccc}\toprule
& & \multicolumn{2}{c}{CGA-PMD} & \multicolumn{2}{c}{DGN-PMD} \\
\cmidrule(lr){3-4}\cmidrule(lr){5-6} $N_2/N_1$ & selection & success & mean & success & mean \\\midrule
1/5 & degree & 10/10 & 4.66 & 10/10 & 4.42 \\
 & random & 10/10 & 4.72 & 10/10 & 3.05 \\
1/3 & degree & 10/10 & 4.69 & 10/10 & 4.76 \\
 & random & 10/10 & 5.00 & 10/10 & 4.35 \\
1/2 & degree & 10/10 & 4.75 & 10/10 & 4.66 \\
 & random & 10/10 & 4.33 & 10/10 & 3.84 \\
\bottomrule\end{tabular}\end{table}

\subsection{Robustness to a misspecified coupling}\label{app:cr_misspec}
The environment keeps the true coupling ($d_{\max}=2$) while the agent's attention mask is built from $\hat\Phi$ with a different radius, at training and at evaluation: the mask is too narrow ($\hat d_{\max}=1$), correct, or too wide ($\hat d_{\max}=3$). ChicagoSketch BFS, 10 seeds, CGA-PMD.

\begin{table}[h]\centering\small
\caption{Coupling misspecification (10 seeds, CGA-PMD): success counts and mean total reward ($\times 10^{10}$).}
\label{tab:cr_misspec}
\begin{tabular}{lccccc}\toprule $N$ & $\hat d_{\max}$ & mean $|\hat\Phi|$ & overlap with $\Phi$ (recall / precision) & success & mean \\\midrule
100 & 1 & 2.1 & 0.35 / 0.95 & 10/10 & 4.66 \\
 & 2 & 5.9 & 1.00 / 1.00 & 10/10 & 4.69 \\
 & 3 & 11.8 & 1.00 / 0.49 & 10/10 & 4.72 \\
\midrule
200 & 1 & 2.1 & 0.30 / 0.91 & 10/10 & 5.45 \\
 & 2 & 6.3 & 1.00 / 1.00 & 10/10 & 6.15 \\
 & 3 & 12.8 & 0.99 / 0.47 & 10/10 & 5.63 \\
\bottomrule\end{tabular}\end{table}

\subsection{Bass-parameter sensitivity}\label{app:cr_pq}
ChicagoSketch BFS at $N=100$, 5 seeds, hydrogen costs held fixed while the innovation and imitation rates are varied by a factor of two in each direction around the calibrated $(p,q) = (0.004, 0.3)$.

\begin{table}[h]\centering\small
\caption{Bass-parameter sweep: mean total reward ($\times 10^{10}$), success counts in parentheses.}
\label{tab:cr_pq}
\begin{tabular}{llccc}\toprule
Method & $p$ & $q=0.15$ & $q=0.3$ & $q=0.6$ \\\midrule
CGA-PMD & 0.002 & 1.10 (5/5) & 3.30 (5/5) & 7.58 (5/5) \\
 & 0.004 & 1.99 (5/5) & 4.68 (5/5) & 8.49 (5/5) \\
 & 0.008 & 3.42 (5/5) & 6.44 (5/5) & 9.97 (5/5) \\
\midrule
IPPO & 0.002 & 0.58 (5/5) & 1.10 (3/5) & 2.02 (3/5) \\
 & 0.004 & 1.49 (4/5) & 1.31 (5/5) & 5.62 (4/5) \\
 & 0.008 & 1.85 (5/5) & 3.81 (4/5) & 6.52 (4/5) \\
\bottomrule\end{tabular}\end{table}

The ordering is unchanged across the sweep: CGA-PMD has the higher mean in 9 of the 9 $(p,q)$ cells.
Faster diffusion raises every method's reward, so the comparison in Section~\ref{sec:experiments} does not rest on the particular adoption curve we calibrate.

\subsection{PPO hyperparameter sweep on ten seeds}\label{app:cr_ppo}
The configurations that matter are reported on the full ten-seed pool at the two scales where CGA-PPO is weakest, against CGA-PMD's cells of Table~\ref{tab:appA_bfs} ($10/10$ with mean $6.17{\cdot}10^{10}$ at $N{=}250$; $9/10$ with $4.13{\cdot}10^{10}$ at $N{=}300$, median training time $48$\,s).

\begin{table}[h]\centering\small
\caption{CGA-PPO configurations on ten seeds: success counts, mean total reward ($\times 10^{10}$) and median training time. CGA-PMD is repeated for reference.}
\label{tab:cr_ppo}
\begin{tabular}{lcccccc}\toprule & \multicolumn{3}{c}{$N{=}250$} & \multicolumn{3}{c}{$N{=}300$} \\
\cmidrule(lr){2-4}\cmidrule(lr){5-7} configuration & success & mean & time (s) & success & mean & time (s) \\\midrule
base & 7/10 & 0.63 & 157 & 3/10 & 0.93 & 136 \\
clip0.1 & 7/10 & 0.63 & 158 & 5/10 & 2.42 & 136 \\
lr1e-3 & 8/10 & 2.96 & 158 & 8/10 & 0.95 & 167 \\
epochs5 & 4/10 & 0.08 & 211 & 8/10 & 2.40 & 208 \\
\midrule CGA-PMD (reference) & 10/10 & 6.17 & 48 & 9/10 & 4.13 & 48 \\
\bottomrule\end{tabular}\end{table}

\subsection{More than two graphs}\label{app:cr_multi}
A toy three-graph instance ($N_1=30$ Barab\'asi--Albert demand graph, two supply graphs): a demand node is operational when it is placed and each supply graph has a placed partner, and the architecture takes one $\Phi$-masked attention block per partner graph with the per-node scores averaged. 5 seeds; nothing else in Algorithm~\ref{alg:cga} changes.

Both variants clear the hydrogen success threshold of $5{\cdot}10^{7}$ on all five seeds, which is what this instance is meant to show: the formalism, the architecture and Algorithm~\ref{alg:cga} carry over to three graphs with no change beyond adding an attention block per partner graph.
The mask makes no measurable difference in mean total reward ($2.75$ vs.\ $2.82{\times}10^{8}$ on an instance far smaller than those of Table~\ref{tab:appA_bfs}), consistent with the two-graph result that the mask separates methods at scale rather than on small instances.

\subsection{Additional baselines}\label{app:cr_grid}
DGN-PPO is CGA-PPO with all-ones coupling matrices, exactly as DGN-PMD is CGA-PMD with all-ones matrices, and MAPPO-Het-PMD keeps MAPPO-Het's unmasked heterogeneous actors and central critic while replacing the MAPPO update with the dual-averaging step of Algorithm~\ref{alg:cga}.
Same instances, same ten seeds, same protocol as Table~\ref{tab:appA_bfs}.

\begin{table}[h]\centering\small
\caption{ChicagoSketch BFS (10 seeds): success counts and mean total reward ($\times 10^{10}$). CGA-PMD from Table~\ref{tab:appA_bfs} is repeated for reference.}
\label{tab:cr_grid}
\resizebox{\textwidth}{!}{%
\begin{tabular}{lcccccc}\toprule Method & $N{=}50$ & $N{=}100$ & $N{=}150$ & $N{=}200$ & $N{=}250$ & $N{=}300$ \\\midrule
DGN-PPO & 10/10, 2.60 & 10/10, 2.28 & 10/10, 3.30 & 10/10, 1.95 & 9/10, 3.36 & 8/10, 2.64 \\
MAPPO-Het-PMD & 10/10, 2.07 & 9/10, 2.00 & 5/10, 2.51 & 7/10, 1.98 & 6/10, 2.20 & 5/10, 0.58 \\
\midrule
CGA-PMD & 10/10 \ 5.92 & 10/10 \ 5.02 & 10/10 \ 5.63 & 10/10 \ 6.51 & 10/10 \ 6.17 & 9/10 \ 4.13 \\
\bottomrule\end{tabular}}
\end{table}

Neither reaches CGA-PMD's cell at any scale. DGN-PPO, unmasked attention without the PMD update, is reliable (8--10 successes of ten) but its mean reward never exceeds 3.36$\times 10^{10}$, roughly half of CGA-PMD's at the same scales. MAPPO-Het-PMD, the PMD update on a central critic without the mask, loses reliability as $N$ grows, from 10/10 to 5/10.

\subsection{Optimality gap on solvable instances}\label{app:cr_exact}
On instances small enough to enumerate, we compare the learned policy with the exact optimum.
Installation costs do not depend on time and adoption is monotone in how early a node is placed, so an optimal schedule never idles before its last placement; it therefore suffices to enumerate, for each agent, an ordered subset of its nodes placed at consecutive steps from $t=0$.
We also sample schedules with gaps as a check, and none beats the enumerated optimum.

\begin{table}[h]\centering\small
\caption{Exact benchmark on reduced instances ($T{=}30$, 5 seeds): mean fraction of the optimum attained.}
\label{tab:cr_exact}
\resizebox{\textwidth}{!}{%
\begin{tabular}{lcccccccc}\toprule
$N_1$ & optimum ($\times10^{10}$) & greedy & IPPO & Indep-PMD & CGA-PPO & CGA-PMD & MAPPO-Het & DGN-PMD \\\midrule
6 & 2.46 & 81\% & 80\% & 88\% & 92\% & 93\% & 87\% & 93\% \\
7 & 2.50 & 81\% & 77\% & 83\% & 81\% & 92\% & 68\% & 92\% \\
\bottomrule\end{tabular}}
\end{table}

CGA-PMD reaches 93\% of the optimum at $N_1=6$ and 92\% at $N_1=7$, tied with DGN-PMD and ahead of every other method at both sizes; every method except IPPO and MAPPO-Het at $N_1=7$ lands within twenty points of the optimum on instances this small. The learned policy is therefore good but not optimal where enumeration is possible, which is the regime the paper targets: what the method buys is reliability and solution quality at the scales where an exact solve is out of reach and the alternatives stop training.

The tie with DGN-PMD reads the same way from the other side: with six or seven demand nodes there is almost no context to filter, so masking it cannot help, and the two separate only as the context grows; DGN-PMD's mean reward falls to $1.59$ and $0.45{\times}10^{10}$ at $N{=}250$ and $N{=}300$ against CGA-PMD's $6.17$ and $4.13$ (Table~\ref{tab:appA_bfs}).

\newpage
\section*{NeurIPS Paper Checklist}

\begin{enumerate}

\item {\bf Claims}
    \item[] Question: Do the main claims made in the abstract and introduction accurately reflect the paper's contributions and scope?
    \item[] Answer: \answerYes{}
    \item[] Justification: The abstract and Introduction state three contributions.

\item {\bf Limitations}
    \item[] Question: Does the paper discuss the limitations of the work performed by the authors?
    \item[] Answer: \answerYes{}
    \item[] Justification: The concluding paragraph of Section~\ref{sec:conclusion} states them.

\item {\bf Theory assumptions and proofs}
    \item[] Question: For each theoretical result, does the paper provide the full set of assumptions and a complete (and correct) proof?
    \item[] Answer: \answerYes{}
    \item[] Justification: Detailed assumptions and complete proofs are shown in Appendix~\ref{app:proof}.

    \item {\bf Experimental result reproducibility}
    \item[] Question: Does the paper fully disclose all the information needed to reproduce the main experimental results of the paper to the extent that it affects the main claims and/or conclusions of the paper (regardless of whether the code and data are provided or not)?
    \item[] Answer: \answerYes{}
    \item[] Justification: Hyperparameters, network construction, coupling definition, environment parameters, and per-baseline implementation details are fully reported in Appendices~\ref{app:experimental} and~\ref{app:reward_params}.

\item {\bf Open access to data and code}
    \item[] Question: Does the paper provide open access to the data and code, with sufficient instructions to faithfully reproduce the main experimental results, as described in supplemental material?
    \item[] Answer: \answerYes{}
    \item[] Justification: All real transportation networks used in the paper (Sioux Falls, Eastern Massachusetts, Friedrichshain, Tiergarten, BerlinMitte, and the ChicagoSketch network used for BFS subsampling in Section~\ref{sec:scaling}) are publicly available from the Transportation Networks repository~\citep{stabler2024transportation}. The CGA-PMD implementation is released at \url{https://github.com/Mina-Kim-Github/cegg-cga-pmd}.

\item {\bf Experimental setting/details}
    \item[] Question: Does the paper specify all the training and test details (e.g., data splits, hyperparameters, how they were chosen, type of optimizer) necessary to understand the results?
    \item[] Answer: \answerYes{}
    \item[] Justification: All training and evaluation details are listed in Appendix~\ref{app:experimental}.     

\item {\bf Experiment statistical significance}
    \item[] Question: Does the paper report error bars suitably and correctly defined or other appropriate information about the statistical significance of the experiments?
    \item[] Answer: \answerYes{}
    \item[] Justification: Main experiments are reported as mean over 10 random seeds per (method, network) cell with 10 greedy rollouts per seed. Standard deviations appear as error bars in Figure~\ref{fig:scaling_meanstd} (BFS mean reward), and per-cell mean$\pm$std with success counts in Tables~\ref{tab:appA_bfs}--\ref{tab:appA_3domain}.    

\item {\bf Experiments compute resources}
    \item[] Question: For each experiment, does the paper provide sufficient information on the computer resources (type of compute workers, memory, time of execution) needed to reproduce the experiments?
    \item[] Answer: \answerYes{}
    \item[] Justification: All experiments run on a single CPU (Apple M-series laptop) with no GPU required. Per-method training times range from roughly 20 seconds (Sioux Falls, $N{=}24$) to ${\sim}17$ minutes per seed for CGA-PPO at $N{\approx}224$; PMD-class methods are roughly $4\times$ faster than CGA-PPO at the same $N$. The full body experimental sweep (Sections~\ref{sec:scaling}--\ref{sec:realnets}) completes in approximately 10 CPU-hours on a single machine; the Figure~\ref{fig:stationarity} cumulative-KL trajectories add ${\sim}1.5$ CPU-hours; an additional cumulative-KL stationarity scan on Friedrichshain $N{=}224$ (3 methods, 10 seeds; not reported in the body) adds ${\sim}4.5$ CPU-hours. The experiments of Appendix~\ref{app:cr} take a further ${\sim}10.5$ CPU-hours in total on the same machine, the largest single items being the two added grid cells (${\sim}2.7$ hours) and the ten-seed PPO sweep at $N{=}250$ (${\sim}2.2$ hours).

\item {\bf Code of ethics}
    \item[] Question: Does the research conducted in the paper conform, in every respect, with the NeurIPS Code of Ethics \url{https://neurips.cc/public/EthicsGuidelines}?
    \item[] Answer: \answerYes{}
    \item[] Justification: The research involves only synthetic graphs and publicly available transportation-network datasets, with no human subjects, personal data, or dual-use concerns. The authors have reviewed and adhered to the NeurIPS Code of Ethics.

\item {\bf Broader impacts}
    \item[] Question: Does the paper discuss both potential positive societal impacts and negative societal impacts of the work performed?
    \item[] Answer: \answerYes{}
    \item[] Justification: The framework is motivated by coordinated planning of low-carbon infrastructure, hydrogen or EV chargers, and its positive societal impact is discussed in the Introduction and Conclusion. We do not foresee direct negative societal impacts, as the method applies to facility-siting decisions rather than to sensitive personal or generative settings.

\item {\bf Safeguards}
    \item[] Question: Does the paper describe safeguards that have been put in place for responsible release of data or models that have a high risk for misuse (e.g., pre-trained language models, image generators, or scraped datasets)?
    \item[] Answer: \answerNA{}
    \item[] Justification: The paper releases neither pre-trained generative models nor scraped datasets, and the proposed infrastructure-planning method has no identified high-risk misuse pathway.

\item {\bf Licenses for existing assets}
    \item[] Question: Are the creators or original owners of assets (e.g., code, data, models), used in the paper, properly credited and are the license and terms of use explicitly mentioned and properly respected?
    \item[] Answer: \answerYes{}
    \item[] Justification: The transportation networks (Sioux Falls, Eastern Massachusetts, Friedrichshain, Tiergarten, BerlinMitte, and ChicagoSketch) come from the Transportation Networks for Research project~\citep{stabler2024transportation} and are cited in the experimental setup. That project makes its data available for academic research and asks that the source be indicated in any publication relying on it, which this paper and the code release both do. All referenced prior methods (IPPO, MAPPO-Het, PMD, PPO, DGN, GraphSAGE, Bass diffusion, etc.) are cited with their original publications.

\item {\bf New assets}
    \item[] Question: Are new assets introduced in the paper well documented and is the documentation provided alongside the assets?
    \item[] Answer: \answerYes{}
    \item[] Justification: The new assets are the CEGG formulation, the CGA architecture, the CGA-PMD training procedure, and the released code. The first three are fully specified in the main text (definitions, equations, Algorithm~\ref{alg:cga}) with reproducibility details in Appendix~\ref{app:experimental}; the implementation is released at \url{https://github.com/Mina-Kim-Github/cegg-cga-pmd}.

\item {\bf Crowdsourcing and research with human subjects}
    \item[] Question: For crowdsourcing experiments and research with human subjects, does the paper include the full text of instructions given to participants and screenshots, if applicable, as well as details about compensation (if any)?
    \item[] Answer: \answerNA{}
    \item[] Justification: The paper involves no crowdsourcing and no research with human subjects.

\item {\bf Institutional review board (IRB) approvals or equivalent for research with human subjects}
    \item[] Question: Does the paper describe potential risks incurred by study participants, whether such risks were disclosed to the subjects, and whether Institutional Review Board (IRB) approvals (or an equivalent approval/review based on the requirements of your country or institution) were obtained?
    \item[] Answer: \answerNA{}
    \item[] Justification: The paper involves no research with human subjects, so no IRB approval was required.

\item {\bf Declaration of LLM usage}
    \item[] Question: Does the paper describe the usage of LLMs if it is an important, original, or non-standard component of the core methods in this research? Note that if the LLM is used only for writing, editing, or formatting purposes and does \emph{not} impact the core methodology, scientific rigor, or originality of the research, declaration is not required.
    \item[] Answer: \answerNA{}
    \item[] Justification: LLMs are not a component of the core methodology of this paper; the proposed CEGG framework and CGA-PMD algorithm do not use LLMs in their formulation, training, or evaluation.

\end{enumerate}

\end{document}